\documentclass[letterpaper]{article} 
\usepackage{aaai2026}  
\usepackage{times}  
\usepackage{helvet}  
\usepackage{courier}  
\usepackage[hyphens]{url}  
\usepackage{graphicx} 
\usepackage{natbib}  
\usepackage{caption} 
\usepackage{multirow}
\usepackage{amsmath,amssymb}
\usepackage{algorithm}
\usepackage{algorithmic}
\usepackage{booktabs}
\usepackage{xcolor}
\usepackage[table]{xcolor}
\usepackage{amsmath}
\usepackage{amssymb}
\title{SpaCRD: Multimodal Deep Fusion of Histology and Spatial Transcriptomics for Cancer Region Detection}
\author {
    Shuailin Xue\textsuperscript{\rm 1,2},
    Jun Wan\textsuperscript{\rm 3},
    Lihua Zhang\textsuperscript{\rm 4},
    Wenwen Min\textsuperscript{\rm 1,2}\thanks{Corresponding author.}
}
\affiliations {
    \textsuperscript{\rm 1}School of Information Science and Engineering, Yunnan University, Kunming 650500, China\\
    \textsuperscript{\rm 2}Yunnan Key Laboratory of Intelligent Systems and Computing, Yunnan University, Kunming 650500, China\\
    \textsuperscript{\rm 3}School of Information Engineering, Zhongnan University of Economics and Law, Wuhan 430073, China\\
    \textsuperscript{\rm 4}School of Artificial Intelligence, School of Computer Science, Wuhan University, Wuhan 430072, China\\
    xueshuailin@stu.ynu.edu.cn, junwan2014@whu.edu.cn, zhanglh@whu.edu.cn, minwenwen@ynu.edu.cn
}

\begin{document}
\maketitle

\begin{abstract}
Accurate detection of cancer tissue regions (CTR) enables deeper analysis of the tumor microenvironment and offers crucial insights into treatment response. Traditional CTR detection methods, which typically rely on the rich cellular morphology in histology images, are susceptible to a high rate of false positives due to morphological similarities across different tissue regions. The groundbreaking advances in spatial transcriptomics (ST) provide detailed cellular phenotypes and spatial localization information, offering new opportunities for more accurate cancer region detection. However, current methods are unable to effectively integrate histology images with ST data, especially in the context of cross-sample and cross-platform/batch settings for accomplishing the CTR detection. To address this challenge, we propose SpaCRD, a transfer learning-based method that deeply integrates histology images and ST data to enable reliable CTR detection across diverse samples, platforms, and batches. Once trained on source data, SpaCRD can be readily generalized to accurately detect cancerous regions across samples from different platforms and batches. The core of SpaCRD is a category-regularized variational reconstruction-guided bidirectional cross-attention fusion network, which enables the model to adaptively capture latent co-expression patterns between histological features and gene expression from multiple perspectives. Extensive benchmark analysis on 23 matched histology-ST datasets spanning various disease types, platforms, and batches demonstrates that SpaCRD consistently outperforms existing eight state-of-the-art methods in CTR detection.
\end{abstract}

\begin{links}
    \link{Code}{https://github.com/wenwenmin/SpaCRD}
\end{links}

\section{Introduction}
In clinical diagnosis and medical research, cancer tissue regions (CTR) detection is a critical step in developing treatment strategies for oncology patients \cite{lawrence2023circulating}. It not only aids in delineating surgical margins and precisely delivering radiation doses, but also provides essential spatial references for tumor microenvironment analysis \cite{khalighi2024artificial,histex}. Prior research has commonly relied on manual annotations from pathologists and traditional anomaly detection algorithms based on histology images analysis for CTR detection \cite{shmatko2022artificial,zingman2024learning}. However, the former is limited by its high cost and time-consuming nature, while the latter suffers from poor accuracy due to the misleading morphological similarities across different tissue regions and inconsistent staining quality in histology images \cite{xue2025inferring}. As a result, neither is an ideal solution for CTR detection in clinical diagnosis and biomedical research.

\begin{figure}[t]
\centering
\includegraphics[width=1\columnwidth]{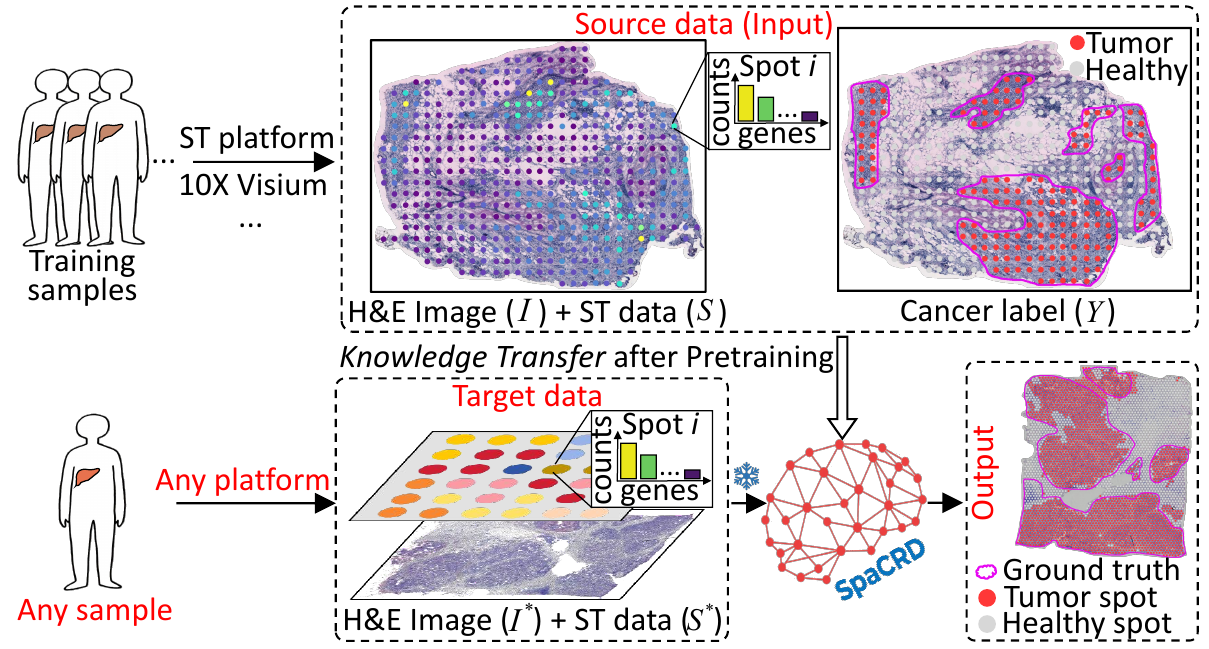}
\caption{SpaCRD achieves accurate CTR detection across diverse samples and platforms\&batches by deeply integrating histology images and ST data through transfer learning.}
\label{fig1}
\end{figure}

As a revolutionary technology, spatial transcriptomics (ST) enables comprehensive profiling of transcripts across entire tissue sections while preserving their spatial locations, offering unprecedented capabilities for characterizing spatial heterogeneity and abnormal tissue architecture \cite{rao2021exploring,tian2023expanding,min2025spavit}. However, to precisely localize sequencing positions, ST requires additional processing steps during sequencing, which inevitably introduce background noise into the gene expression \cite{janssen2023effect}. ST-based algorithms \cite{benjamin2024multiscale,ferri2023spatial,shen2022spatial} for various tasks are inevitably affected by the substantial noise inherent in ST data, which often prevents them from achieving optimal performance \cite{zahedi2024deep}.

In addition, current tissue annotation methods \cite{istar,tesla}, which rely heavily on expert-defined prior knowledge, also detect CTR by aggregating the expression of marker genes. However, identifying reliable markers often requires extensive domain knowledge and experimental validation, and many cancer types lack well-defined markers that have been systematically identified, which further limits the generalizability and applicability of such approaches.

Effectively integrating histology and ST data to overcome morphological ambiguities and ST noise remains a major challenge in CTR detection \cite{maan2025multi}. While some multimodal methods have been proposed, they suffer from critical shortcomings. For instance, SpaCell \cite{tan2020spacell} fuses histology and ST data through simple feature concatenation, neglecting cross-modal interactions and global spatial context. STANDS \cite{xu2024detecting} and MEATRD \cite{xu2025meatrd} follow the paradigm of traditional visual anomaly detection \cite{liu2023simplenet}, but their reliance on reconstruction errors is ill-suited for structured cancer regions, which differ fundamentally from sparse anomalies in natural images \cite{seferbekova2023spatial}. Furthermore, these methods often fail to generalize across datasets due to batch heterogeneity, highlighting the need for more robust solutions.

Transfer learning offers a promising solution by leveraging knowledge learned from well-annotated source datasets to improve performance on heterogeneous target domains \cite{huangscalable,xue2024stentrans}. For example, several studies have used single-cell RNA-seq datasets from similar tissues as the source domain to guide cell type classification in ST datasets (target domain), effectively transferring cell-type signatures to spatial contexts \cite{yan2025triple,hao2021integrated}. In CTR detection tasks, the target samples often originate from different platforms and experimental batches, resulting in substantial technical and biological variability. Motivated by the success of transfer learning, we apply transfer learning to align heterogeneous ST datasets and improve the generalizability of CTR detection across platforms and batches.

In this study, we propose \textbf{SpaCRD}, a multimodal deep fusion framework that integrates histology images and \textbf{Spa}tial transcriptomics data for \textbf{C}ancer \textbf{R}egion \textbf{D}etection. By leveraging transfer learning and a powerful multimodal deep fusion module, SpaCRD effectively mitigates technical and batch variations among affected individuals, achieving accurate and consistent CTR detection performance across different samples, platforms, and batches. Specifically, SpaCRD comprises a pretrained pathology foundation model, a modality-alignment representation learning, and a category-regularized \textbf{V}ariational \textbf{R}econstruction-guided \textbf{B}idirectional \textbf{C}ross-\textbf{A}ttention fusion network (VRBCA), which collectively enable end-to-end CTR detection from multimodal inputs. To the best of our knowledge, SpaCRD is the first framework that combines multimodal deep fusion with transfer learning for CTR detection. In summary, our contributions can be summarized as follows:
\begin{itemize}
    \item We propose SpaCRD, a novel CTR detection framework leveraging multimodal deep fusion and transfer learning.
    \item SpaCRD mitigates technical and batch effects by aligning samples of the same disease type into a consistent representation space, thus enabling CTR detection across diverse samples and platforms\&batches.
    \item We design a VRBCA network that reduces modality discrepancies, filters out noise, and stabilizes the fusion of ST data and histology images from complementary perspectives, while comprehensively modeling interactions among neighboring spots to facilitate the generation of compact and class-specific multimodal embeddings.
    \item Extensive benchmark on eleven breast cancer and twelve colorectal cancer datasets demonstrate that SpaCRD consistently outperforms eight state-of-the-art (SOTA) methods in CTR detection.
\end{itemize}

\begin{figure*}[t]
\centering
\includegraphics[width=1\textwidth]{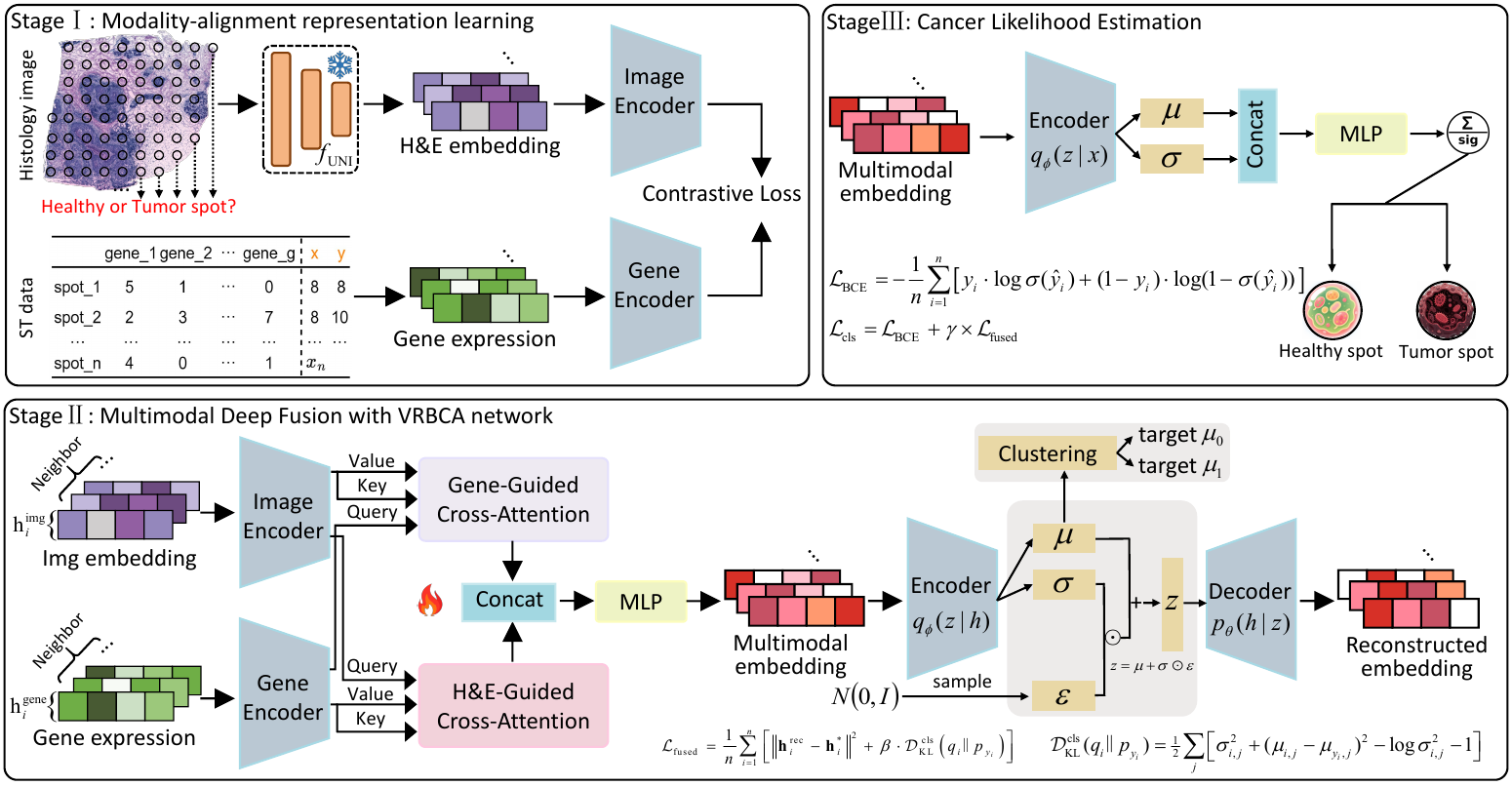}
\caption{The framework of SpaCRD. Stage I: UNI is used to extract histology features, while modality-alignment representation learning aligns histology and ST modalities into a shared embedding space. Stage II: VRBCA integrates aligned histology and ST features into a compact and class-consistent embedding that captures biologically relevant cross-modal interactions. Stage III: The learned representation is used to estimate cancer likelihood scores for each spot.}
\label{fig2}
\end{figure*}

\begin{table*}[t]
\centering
\resizebox{1\textwidth}{!}{
\begin{tabular}{l|l|cccccccccccc}
\toprule
\multirow{2}{*}{\textbf{Metric}} & \multirow{2}{*}{\textbf{Method}} &\multicolumn{11}{c}{\textbf{Cross-samples}}  \\
& & CRC\_A1& CRC\_A2& CRC\_B1& CRC\_B2& CRC\_C1& CRC\_C2& CRC\_D1& CRC\_D2& CRC\_E1& CRC\_E2& CRC\_G1 & CRC\_G2\\
\hline
\multirow{9}{*}{\textbf{AUC}}&SimpleNet& 0.491\scriptsize{$\pm$0.163} & 0.518\scriptsize{$\pm$0.098} & 0.488\scriptsize{$\pm$0.036} & 0.481\scriptsize{$\pm$0.055} & 0.440\scriptsize{$\pm$0.047} & 0.488\scriptsize{$\pm$0.070} & 0.518\scriptsize{$\pm$0.050} & 0.435\scriptsize{$\pm$0.036} & 0.429\scriptsize{$\pm$0.129} & 0.580\scriptsize{$\pm$0.103} & 0.506\scriptsize{$\pm$0.082} & 0.463\scriptsize{$\pm$0.047} \\
&Spatial-ID & 0.594\scriptsize{$\pm$0.068} & 0.563\scriptsize{$\pm$0.091} & 0.518\scriptsize{$\pm$0.016} & 0.496\scriptsize{$\pm$0.043} & 0.513\scriptsize{$\pm$0.057} & 0.503\scriptsize{$\pm$0.082} & 0.510\scriptsize{$\pm$0.029} & 0.526\scriptsize{$\pm$0.040} & 0.457\scriptsize{$\pm$0.109} & 0.453\scriptsize{$\pm$0.093} & 0.458\scriptsize{$\pm$0.042} & 0.525\scriptsize{$\pm$0.033} \\
&STAGE& 0.544\scriptsize{$\pm$0.088}& 0.516\scriptsize{$\pm$0.116}& 0.517\scriptsize{$\pm$0.037}& 0.520\scriptsize{$\pm$0.078}& 0.488\scriptsize{$\pm$0.084}& 0.502\scriptsize{$\pm$0.059}& 0.498\scriptsize{$\pm$0.086}& 0.490\scriptsize{$\pm$0.055}& 0.354\scriptsize{$\pm$0.129}& 0.466\scriptsize{$\pm$0.086}& 0.488\scriptsize{$\pm$0.106}& 0.537\scriptsize{$\pm$0.117}\\
&STANDS& 0.540\scriptsize{$\pm$0.045}& 0.530\scriptsize{$\pm$0.043}& 0.466\scriptsize{$\pm$0.030}& 0.506\scriptsize{$\pm$0.021}& 0.587\scriptsize{$\pm$0.053}& 0.549\scriptsize{$\pm$0.057}& 0.564\scriptsize{$\pm$0.053}& 0.510\scriptsize{$\pm$0.022}& 0.609\scriptsize{$\pm$0.075}& 0.643\scriptsize{$\pm$0.098}& 0.504\scriptsize{$\pm$0.027}& 0.600\scriptsize{$\pm$0.070}\\
&TESLA& 0.496\scriptsize{$\pm$0.016}& 0.527\scriptsize{$\pm$0.004}& 0.436\scriptsize{$\pm$0.006}& 0.548\scriptsize{$\pm$0.007}& 0.716\scriptsize{$\pm$0.011}& 0.593\scriptsize{$\pm$0.002}& \underline{0.664}\scriptsize{$\pm$0.006}& 0.582\scriptsize{$\pm$0.036}& 0.517\scriptsize{$\pm$0.007}& 0.635\scriptsize{$\pm$0.024}& 0.662\scriptsize{$\pm$0.004}& 0.721\scriptsize{$\pm$0.006}\\
&iStar& 0.444\scriptsize{$\pm$0.025}& 0.436\scriptsize{$\pm$0.027}& 0.517\scriptsize{$\pm$0.002}& 0.514\scriptsize{$\pm$0.005}& 0.733\scriptsize{$\pm$0.004}& 0.638\scriptsize{$\pm$0.004}& 0.654\scriptsize{$\pm$0.008}& \textbf{0.672}\scriptsize{$\pm$0.061}& 0.505\scriptsize{$\pm$0.014}& 0.593\scriptsize{$\pm$0.051}& \underline{0.781}\scriptsize{$\pm$0.002}& \underline{0.799}\scriptsize{$\pm$0.005}\\
&MEATRD& 0.505\scriptsize{$\pm$0.007}& 0.517\scriptsize{$\pm$0.008}& 0.511\scriptsize{$\pm$0.005}& 0.548\scriptsize{$\pm$0.002}& \underline{0.765}\scriptsize{$\pm$0.008}& 0.539\scriptsize{$\pm$0.009}& 0.498\scriptsize{$\pm$0.007}& 0.466\scriptsize{$\pm$0.013}& 0.541\scriptsize{$\pm$0.005}& \underline{0.904}\scriptsize{$\pm$0.011}& 0.507\scriptsize{$\pm$0.006}& 0.550\scriptsize{$\pm$0.006}\\
&SpaCell-Plus& \underline{0.821}\scriptsize{$\pm$0.012}& \underline{0.762}\scriptsize{$\pm$0.010}& \underline{0.678}\scriptsize{$\pm$0.006}& \underline{0.713}\scriptsize{$\pm$0.004}& 0.761\scriptsize{$\pm$0.014}& \underline{0.713}\scriptsize{$\pm$0.012}& 0.625\scriptsize{$\pm$0.034}& 0.472\scriptsize{$\pm$0.018}& \underline{0.799}\scriptsize{$\pm$0.024}& 0.816\scriptsize{$\pm$0.015}& 0.638\scriptsize{$\pm$0.028}& 0.733\scriptsize{$\pm$0.016}\\
&SpaCRD(ours)& \textbf{0.953}\scriptsize{$\pm$0.002}& \textbf{0.925}\scriptsize{$\pm$0.004}& \textbf{0.824}\scriptsize{$\pm$0.005}& \textbf{0.897}\scriptsize{$\pm$0.003}& \textbf{0.895}\scriptsize{$\pm$0.007}& \textbf{0.880}\scriptsize{$\pm$0.010}& \textbf{0.789}\scriptsize{$\pm$0.020}& \underline{0.603}\scriptsize{$\pm$0.022}& \textbf{0.966}\scriptsize{$\pm$0.006}& \textbf{0.961}\scriptsize{$\pm$0.007}& \textbf{0.853}\scriptsize{$\pm$0.005}& \textbf{0.888}\scriptsize{$\pm$0.003}\\
\hline
\multirow{9}{*}{\textbf{AP}}&SimpleNet& 0.312\scriptsize{$\pm$0.079} & 0.356\scriptsize{$\pm$0.070} & 0.688\scriptsize{$\pm$0.027} & 0.703\scriptsize{$\pm$0.039} & 0.239\scriptsize{$\pm$0.019} & 0.425\scriptsize{$\pm$0.042} & 0.869\scriptsize{$\pm$0.016} & 0.888\scriptsize{$\pm$0.010} & 0.342\scriptsize{$\pm$0.067} & 0.598\scriptsize{$\pm$0.081} & 0.165\scriptsize{$\pm$0.034} & 0.188\scriptsize{$\pm$0.015} \\
&Spatial-ID & 0.366\scriptsize{$\pm$0.023} & 0.401\scriptsize{$\pm$0.103} & 0.707\scriptsize{$\pm$0.019} & 0.712\scriptsize{$\pm$0.025} & 0.285\scriptsize{$\pm$0.045} & 0.441\scriptsize{$\pm$0.070} & 0.866\scriptsize{$\pm$0.012} & 0.912\scriptsize{$\pm$0.012} & 0.352\scriptsize{$\pm$0.085} & 0.512\scriptsize{$\pm$0.059} & 0.141\scriptsize{$\pm$0.011} & 0.215\scriptsize{$\pm$0.023} \\
&STAGE& 0.300\scriptsize{$\pm$0.054}& 0.335\scriptsize{$\pm$0.092}& 0.686\scriptsize{$\pm$0.025}& 0.712\scriptsize{$\pm$0.044}& 0.257\scriptsize{$\pm$0.047}& 0.419\scriptsize{$\pm$0.043}& 0.852\scriptsize{$\pm$0.031}& \underline{0.913}\scriptsize{$\pm$0.010}& 0.295\scriptsize{$\pm$0.052}& 0.504\scriptsize{$\pm$0.064}& 0.164\scriptsize{$\pm$0.045}& 0.240\scriptsize{$\pm$0.089}\\
&STANDS& 0.319\scriptsize{$\pm$0.038}& 0.351\scriptsize{$\pm$0.039}& 0.664\scriptsize{$\pm$0.020}& 0.713\scriptsize{$\pm$0.017}& 0.349\scriptsize{$\pm$0.056}& 0.452\scriptsize{$\pm$0.053}& 0.888\scriptsize{$\pm$0.016}& 0.911\scriptsize{$\pm$0.005}& 0.475\scriptsize{$\pm$0.074}& 0.676\scriptsize{$\pm$0.087}& 0.167\scriptsize{$\pm$0.019}& 0.305\scriptsize{$\pm$0.071}\\
&TESLA& 0.317\scriptsize{$\pm$0.008}& 0.421\scriptsize{$\pm$0.024}& 0.574\scriptsize{$\pm$0.004}& 0.613\scriptsize{$\pm$0.005}& 0.376\scriptsize{$\pm$0.004}& 0.536\scriptsize{$\pm$0.006}& 0.874\scriptsize{$\pm$0.003}& 0.813\scriptsize{$\pm$0.017}& 0.427\scriptsize{$\pm$0.004}& 0.642\scriptsize{$\pm$0.022}& 0.425\scriptsize{$\pm$0.002}& 0.360\scriptsize{$\pm$0.004}\\
&iStar& 0.252\scriptsize{$\pm$0.007}& 0.290\scriptsize{$\pm$0.008}& 0.695\scriptsize{$\pm$0.002}& 0.717\scriptsize{$\pm$0.003}& 0.410\scriptsize{$\pm$0.002}& 0.503\scriptsize{$\pm$0.003}& \underline{0.900}\scriptsize{$\pm$0.002}& 0.911\scriptsize{$\pm$0.010}& 0.372\scriptsize{$\pm$0.006}& 0.597\scriptsize{$\pm$0.033}& 0.349\scriptsize{$\pm$0.007}& 0.433\scriptsize{$\pm$0.007}\\
&MEATRD& 0.272\scriptsize{$\pm$0.005}& 0.329\scriptsize{$\pm$0.005}& 0.693\scriptsize{$\pm$0.003}& \underline{0.742}\scriptsize{$\pm$0.005}& 0.518\scriptsize{$\pm$0.023}& 0.439\scriptsize{$\pm$0.007}& 0.861\scriptsize{$\pm$0.005}& 0.902\scriptsize{$\pm$0.002}& 0.403\scriptsize{$\pm$0.005}& \underline{0.892}\scriptsize{$\pm$0.015}& 0.194\scriptsize{$\pm$0.003}& 0.236\scriptsize{$\pm$0.004}\\
&SpaCell-Plus& \underline{0.714}\scriptsize{$\pm$0.035}& \underline{0.695}\scriptsize{$\pm$0.016}& \underline{0.731}\scriptsize{$\pm$0.008}& 0.738\scriptsize{$\pm$0.018}& \underline{0.577}\scriptsize{$\pm$0.020}& \underline{0.670}\scriptsize{$\pm$0.005}& 0.773\scriptsize{$\pm$0.026}& 0.836\scriptsize{$\pm$0.022}& \underline{0.748}\scriptsize{$\pm$0.009}& 0.797\scriptsize{$\pm$0.023}& \underline{0.520}\scriptsize{$\pm$0.036}& \underline{0.624}\scriptsize{$\pm$0.011}\\
&SpaCRD(ours)& \textbf{0.856}\scriptsize{$\pm$0.009}& \textbf{0.853}\scriptsize{$\pm$0.010}& \textbf{0.904}\scriptsize{$\pm$0.014}& \textbf{0.951}\scriptsize{$\pm$0.001}& \textbf{0.749}\scriptsize{$\pm$0.018}& \textbf{0.815}\scriptsize{$\pm$0.026}& \textbf{0.951}\scriptsize{$\pm$0.010}& \textbf{0.936}\scriptsize{$\pm$0.006}& \textbf{0.935}\scriptsize{$\pm$0.008}& \textbf{0.954}\scriptsize{$\pm$0.014}& \textbf{0.619}\scriptsize{$\pm$0.030}& \textbf{0.730}\scriptsize{$\pm$0.005}\\
\hline
\multirow{9}{*}{\textbf{F1}}&SimpleNet& 0.290\scriptsize{$\pm$0.138} & 0.351\scriptsize{$\pm$0.061} & 0.678\scriptsize{$\pm$0.012} & 0.701\scriptsize{$\pm$0.022} & 0.210\scriptsize{$\pm$0.031} & 0.412\scriptsize{$\pm$0.057} & 0.865\scriptsize{$\pm$0.007} & 0.910\scriptsize{$\pm$0.002} & 0.306\scriptsize{$\pm$0.112} & 0.606\scriptsize{$\pm$0.059} & 0.151\scriptsize{$\pm$0.061} & 0.166\scriptsize{$\pm$0.028} \\
&Spatial-ID & 0.359\scriptsize{$\pm$0.044} & 0.364\scriptsize{$\pm$0.086} & 0.688\scriptsize{$\pm$0.005} & 0.705\scriptsize{$\pm$0.026} & 0.283\scriptsize{$\pm$0.066} & 0.418\scriptsize{$\pm$0.064} & 0.864\scriptsize{$\pm$0.006} & 0.918\scriptsize{$\pm$0.002} & 0.326\scriptsize{$\pm$0.123} & 0.510\scriptsize{$\pm$0.078} & 0.116\scriptsize{$\pm$0.019} & 0.217\scriptsize{$\pm$0.049} \\
&STAGE& 0.294\scriptsize{$\pm$0.093}& 0.327\scriptsize{$\pm$0.109}& \underline{0.721}\scriptsize{$\pm$0.016}& \underline{0.726}\scriptsize{$\pm$0.031}& 0.226\scriptsize{$\pm$0.084}& 0.417\scriptsize{$\pm$0.045}& 0.816\scriptsize{$\pm$0.011}& 0.911\scriptsize{$\pm$0.002}& 0.242\scriptsize{$\pm$0.107}& 0.520\scriptsize{$\pm$0.054}& 0.150\scriptsize{$\pm$0.080}& 0.234\scriptsize{$\pm$0.116}\\
&STANDS& 0.300\scriptsize{$\pm$0.048}& 0.344\scriptsize{$\pm$0.041}& 0.671\scriptsize{$\pm$0.011}& 0.711\scriptsize{$\pm$0.010}& 0.326\scriptsize{$\pm$0.064}& 0.465\scriptsize{$\pm$0.043}& \underline{0.866}\scriptsize{$\pm$0.003}& \underline{0.915}\scriptsize{$\pm$0.002}& 0.466\scriptsize{$\pm$0.071}& 0.638\scriptsize{$\pm$0.081}& 0.160\scriptsize{$\pm$0.031}& 0.298\scriptsize{$\pm$0.079}\\
&TESLA& 0.412\scriptsize{$\pm$0.010}& 0.345\scriptsize{$\pm$0.015}& 0.427\scriptsize{$\pm$0.022}& 0.106\scriptsize{$\pm$0.021}& 0.513\scriptsize{$\pm$0.007}& \underline{0.612}\scriptsize{$\pm$0.004}& 0.853\scriptsize{$\pm$0.002}& 0.809\scriptsize{$\pm$0.029}& 0.546\scriptsize{$\pm$0.030}& 0.594\scriptsize{$\pm$0.033}& 0.384\scriptsize{$\pm$0.006}& 0.528\scriptsize{$\pm$0.018}\\
&iStar& 0.332\scriptsize{$\pm$0.037}& 0.318\scriptsize{$\pm$0.011}& 0.103\scriptsize{$\pm$0.027}& 0.058\scriptsize{$\pm$0.019}& 0.576\scriptsize{$\pm$0.003}& 0.667\scriptsize{$\pm$0.003}& 0.872\scriptsize{$\pm$0.005}& 0.744\scriptsize{$\pm$0.150}& 0.434\scriptsize{$\pm$0.026}& 0.572\scriptsize{$\pm$0.092}& 0.531\scriptsize{$\pm$0.009}& \underline{0.608}\scriptsize{$\pm$0.006}\\
&MEATRD& 0.270\scriptsize{$\pm$0.011}& 0.320\scriptsize{$\pm$0.010}& 0.690\scriptsize{$\pm$0.004}& 0.723\scriptsize{$\pm$0.002}& 0.526\scriptsize{$\pm$0.017}& 0.452\scriptsize{$\pm$0.014}& 0.866\scriptsize{$\pm$0.002}& 0.911\scriptsize{$\pm$0.001}& 0.406\scriptsize{$\pm$0.012}& \underline{0.851}\scriptsize{$\pm$0.016}& 0.156\scriptsize{$\pm$0.011}& 0.239\scriptsize{$\pm$0.013}\\
&SpaCell-Plus& \underline{0.592}\scriptsize{$\pm$0.021}& \underline{0.633}\scriptsize{$\pm$0.008}& 0.714\scriptsize{$\pm$0.020}& 0.696\scriptsize{$\pm$0.006}& \underline{0.598}\scriptsize{$\pm$0.016}& 0.573\scriptsize{$\pm$0.012}& 0.726\scriptsize{$\pm$0.036}& 0.823\scriptsize{$\pm$0.008}& \underline{0.710}\scriptsize{$\pm$0.025}& 0.743\scriptsize{$\pm$0.016}& \underline{0.516}\scriptsize{$\pm$0.016}& 0.576\scriptsize{$\pm$0.021}\\
&SpaCRD(ours)& \textbf{0.802}\scriptsize{$\pm$0.006}& \textbf{0.805}\scriptsize{$\pm$0.009}& \textbf{0.836}\scriptsize{$\pm$0.013}& \textbf{0.881}\scriptsize{$\pm$0.005}& \textbf{0.683}\scriptsize{$\pm$0.025}& \textbf{0.777}\scriptsize{$\pm$0.010}& \textbf{0.911}\scriptsize{$\pm$0.002}& \textbf{0.923}\scriptsize{$\pm$0.001}& \textbf{0.894}\scriptsize{$\pm$0.009}& \textbf{0.919}\scriptsize{$\pm$0.006}& \textbf{0.599}\scriptsize{$\pm$0.015}& \textbf{0.694}\scriptsize{$\pm$0.007}\\
\bottomrule
\end{tabular}}
\caption{Quantitative evaluation of SpaCRD for CTR detection on twelve colorectal cancer datasets compared to baselines. Each reported values are means ± standard deviations over five independent runs. Best results in bold, second-best underlined.}
\label{table1}
\end{table*}

\begin{table*}[!ht]
\centering
\resizebox{1\textwidth}{!}{
\begin{tabular}{l|l|cccccccc|ccc}
\toprule
\multirow{2}{*}{\textbf{Metric}} & \multirow{2}{*}{\textbf{Method}} &\multicolumn{8}{c|}{\textbf{Cross-samples}}&\multicolumn{3}{c}{\textbf{Cross-platforms\&batches}}  \\
& & STHBC\_A& STHBC\_B& STHBC\_C& STHBC\_D& STHBC\_E& STHBC\_F& STHBC\_G& STHBC\_H& ViHBC& XeHBC& IDC\\
\hline
\multirow{9}{*}{\textbf{AUC}}&SimpleNet& 0.536\scriptsize{$\pm$0.126}& 0.468\scriptsize{$\pm$0.157}& 0.476\scriptsize{$\pm$0.132}& 0.528\scriptsize{$\pm$0.143}& 0.429\scriptsize{$\pm$0.122}& 0.536\scriptsize{$\pm$0.063}& 0.571\scriptsize{$\pm$0.101}& 0.454\scriptsize{$\pm$0.132}& 0.468\scriptsize{$\pm$0.076}& 0.531\scriptsize{$\pm$0.106}& 0.612\scriptsize{$\pm$0.098} \\
&Spatial-ID & 0.544\scriptsize{$\pm$0.045}& 0.534\scriptsize{$\pm$0.135}& 0.471\scriptsize{$\pm$0.083}& 0.451\scriptsize{$\pm$0.085}& 0.477\scriptsize{$\pm$0.102}& 0.496\scriptsize{$\pm$0.107}& 0.558\scriptsize{$\pm$0.094}& 0.438\scriptsize{$\pm$0.117}& 0.436\scriptsize{$\pm$0.067}& 0.593\scriptsize{$\pm$0.081}& 0.502\scriptsize{$\pm$0.080} \\
&STAGE& 0.536\scriptsize{$\pm$0.126}& 0.468\scriptsize{$\pm$0.157}& 0.476\scriptsize{$\pm$0.132}& 0.528\scriptsize{$\pm$0.143}& 0.429\scriptsize{$\pm$0.122}& 0.536\scriptsize{$\pm$0.063}& 0.571\scriptsize{$\pm$0.101}& 0.474\scriptsize{$\pm$0.190}& 0.444\scriptsize{$\pm$0.067}& 0.575\scriptsize{$\pm$0.054}& 0.605\scriptsize{$\pm$0.068}\\
&STANDS& 0.535\scriptsize{$\pm$0.053}& 0.449\scriptsize{$\pm$0.058}& 0.510\scriptsize{$\pm$0.008}& 0.507\scriptsize{$\pm$0.038}& 0.524\scriptsize{$\pm$0.027}& 0.575\scriptsize{$\pm$0.036}& 0.564\scriptsize{$\pm$0.046}& 0.451\scriptsize{$\pm$0.072}& 0.629\scriptsize{$\pm$0.102}& 0.532\scriptsize{$\pm$0.037}& 0.466\scriptsize{$\pm$0.092}\\
&TESLA& 0.724\scriptsize{$\pm$0.008}& 0.586\scriptsize{$\pm$0.018}& 0.551\scriptsize{$\pm$0.012}& 0.605\scriptsize{$\pm$0.011}& 0.627\scriptsize{$\pm$0.002}& 0.705\scriptsize{$\pm$0.022}& 0.697\scriptsize{$\pm$0.014}& 0.729\scriptsize{$\pm$0.009}& 0.672\scriptsize{$\pm$0.025}& 0.748\scriptsize{$\pm$0.026}& 0.576\scriptsize{$\pm$0.021}\\
&iStar& 0.788\scriptsize{$\pm$0.013}& 0.674\scriptsize{$\pm$0.020}& 0.591\scriptsize{$\pm$0.017}& 0.668\scriptsize{$\pm$0.016}& \underline{0.741}\scriptsize{$\pm$0.004}& 0.672\scriptsize{$\pm$0.024}& 0.714\scriptsize{$\pm$0.017}& 0.681\scriptsize{$\pm$0.011}& 0.736\scriptsize{$\pm$0.032}& \underline{0.842}\scriptsize{$\pm$0.038}& 0.531\scriptsize{$\pm$0.009}\\
&MEATRD& 0.604\scriptsize{$\pm$0.007}& \underline{0.963}\scriptsize{$\pm$0.003}& \underline{0.783}\scriptsize{$\pm$0.010}& 0.837\scriptsize{$\pm$0.006}& 0.442\scriptsize{$\pm$0.013}& 0.468\scriptsize{$\pm$0.023}& 0.774\scriptsize{$\pm$0.029}& 0.614\scriptsize{$\pm$0.024}& 0.517\scriptsize{$\pm$0.002}& 0.472\scriptsize{$\pm$0.004}& 0.496\scriptsize{$\pm$0.002}\\
&SpaCell-Plus& \underline{0.929}\scriptsize{$\pm$0.010}& 0.950\scriptsize{$\pm$0.012}& 0.773\scriptsize{$\pm$0.019}& \underline{0.844}\scriptsize{$\pm$0.009}& 0.668\scriptsize{$\pm$0.056}& \underline{0.774}\scriptsize{$\pm$0.056}& \underline{0.826}\scriptsize{$\pm$0.021}& \underline{0.812}\scriptsize{$\pm$0.017}& \underline{0.784}\scriptsize{$\pm$0.028}& 0.818\scriptsize{$\pm$0.034}& \underline{0.803}\scriptsize{$\pm$0.009}\\
&SpaCRD(ours)& \textbf{0.979}\scriptsize{$\pm$0.002}& \textbf{0.993}\scriptsize{$\pm$0.002}& \textbf{0.795}\scriptsize{$\pm$0.025}& \textbf{0.952}\scriptsize{$\pm$0.003}& \textbf{0.909}\scriptsize{$\pm$0.011}& \textbf{0.913}\scriptsize{$\pm$0.008}& \textbf{0.923}\scriptsize{$\pm$0.004}& \textbf{0.969}\scriptsize{$\pm$0.004}& \textbf{0.900}\scriptsize{$\pm$0.029}& \textbf{0.931}\scriptsize{$\pm$0.008}& \textbf{0.891}\scriptsize{$\pm$0.006}\\
\hline
\multirow{9}{*}{\textbf{AP}}&SimpleNet& 0.907\scriptsize{$\pm$0.038}& 0.226\scriptsize{$\pm$0.062}& 0.724\scriptsize{$\pm$0.070}& 0.511\scriptsize{$\pm$0.097}& 0.536\scriptsize{$\pm$0.083}& 0.870\scriptsize{$\pm$0.024}& 0.412\scriptsize{$\pm$0.082}& 0.287\scriptsize{$\pm$0.062}& 0.603\scriptsize{$\pm$0.047}& 0.570\scriptsize{$\pm$0.054}& 0.662\scriptsize{$\pm$0.073} \\
&Spatial-ID & 0.909\scriptsize{$\pm$0.017}& 0.264\scriptsize{$\pm$0.122}& 0.729\scriptsize{$\pm$0.038}& 0.453\scriptsize{$\pm$0.047}& 0.565\scriptsize{$\pm$0.076}& 0.855\scriptsize{$\pm$0.042}& 0.383\scriptsize{$\pm$0.075}& 0.285\scriptsize{$\pm$0.036}& 0.571\scriptsize{$\pm$0.051}& 0.405\scriptsize{$\pm$0.035}& 0.582\scriptsize{$\pm$0.044} \\
&STAGE& 0.907\scriptsize{$\pm$0.038}& 0.226\scriptsize{$\pm$0.062}& 0.724\scriptsize{$\pm$0.070}& 0.511\scriptsize{$\pm$0.097}& 0.536\scriptsize{$\pm$0.083}& 0.870\scriptsize{$\pm$0.024}& 0.412\scriptsize{$\pm$0.082}& 0.309\scriptsize{$\pm$0.155}&  0.559\scriptsize{$\pm$0.039}& 0.529\scriptsize{$\pm$0.047}& 0.643\scriptsize{$\pm$0.053}\\
&STANDS& 0.889\scriptsize{$\pm$0.019}& 0.179\scriptsize{$\pm$0.018}& 0.731\scriptsize{$\pm$0.023}& 0.451\scriptsize{$\pm$0.037}& 0.557\scriptsize{$\pm$0.024}& 0.859\scriptsize{$\pm$0.017}& 0.345\scriptsize{$\pm$0.029}& 0.290\scriptsize{$\pm$0.016}&  0.692\scriptsize{$\pm$0.094}& 0.575\scriptsize{$\pm$0.060}& 0.573\scriptsize{$\pm$0.070}\\
&TESLA& 0.842\scriptsize{$\pm$0.004}& 0.354\scriptsize{$\pm$0.012}& 0.803\scriptsize{$\pm$0.004}& 0.657\scriptsize{$\pm$0.006}& 0.683\scriptsize{$\pm$0.008}& 0.829\scriptsize{$\pm$0.006}& 0.576\scriptsize{$\pm$0.016}& 0.568\scriptsize{$\pm$0.004}& 0.754\scriptsize{$\pm$0.028}& 0.620\scriptsize{$\pm$0.012}& 0.697\scriptsize{$\pm$0.007}\\
&iStar& 0.956\scriptsize{$\pm$0.003}& 0.332\scriptsize{$\pm$0.020}& 0.783\scriptsize{$\pm$0.009}& 0.632\scriptsize{$\pm$0.011}& \underline{0.761}\scriptsize{$\pm$0.004}& 0.903\scriptsize{$\pm$0.007}& 0.543\scriptsize{$\pm$0.029}& 0.460\scriptsize{$\pm$0.005}& 0.788\scriptsize{$\pm$0.021}& 0.723\scriptsize{$\pm$0.014}& 0.621\scriptsize{$\pm$0.007}\\
&MEATRD& 0.916\scriptsize{$\pm$0.017}& \underline{0.948}\scriptsize{$\pm$0.007}& \underline{0.916}\scriptsize{$\pm$0.005}& 0.832\scriptsize{$\pm$0.013}& 0.554\scriptsize{$\pm$0.010}& 0.848\scriptsize{$\pm$0.008}& 0.653\scriptsize{$\pm$0.024}& 0.459\scriptsize{$\pm$0.011}& 0.631\scriptsize{$\pm$0.005}& 0.317\scriptsize{$\pm$0.008}& 0.595\scriptsize{$\pm$0.003}\\
&SpaCell-Plus& \underline{0.991}\scriptsize{$\pm$0.002}& 0.853\scriptsize{$\pm$0.036}& 0.904\scriptsize{$\pm$0.007}& \underline{0.845}\scriptsize{$\pm$0.011}& 0.757\scriptsize{$\pm$0.043}& \underline{0.949}\scriptsize{$\pm$0.017}& \underline{0.745}\scriptsize{$\pm$0.049}& \underline{0.745}\scriptsize{$\pm$0.013}& \underline{0.846}\scriptsize{$\pm$0.015}& \underline{0.726}\scriptsize{$\pm$0.018}& \underline{0.847}\scriptsize{$\pm$0.006}\\
&SpaCRD(ours)& \textbf{0.998}\scriptsize{$\pm$0.000}& \textbf{0.980}\scriptsize{$\pm$0.003}& \textbf{0.923}\scriptsize{$\pm$0.011}& \textbf{0.961}\scriptsize{$\pm$0.001}& \textbf{0.911}\scriptsize{$\pm$0.013}& \textbf{0.983}\scriptsize{$\pm$0.003}& \textbf{0.867}\scriptsize{$\pm$0.007}& \textbf{0.947}\scriptsize{$\pm$0.005}& \textbf{0.930}\scriptsize{$\pm$0.037}& \textbf{0.859}\scriptsize{$\pm$0.021}& \textbf{0.914}\scriptsize{$\pm$0.006}\\
\hline
\multirow{9}{*}{\textbf{F1}}&SimpleNet& 0.903\scriptsize{$\pm$0.006}& 0.197\scriptsize{$\pm$0.127}& 0.738\scriptsize{$\pm$0.046}& 0.499\scriptsize{$\pm$0.103}& 0.535\scriptsize{$\pm$0.067}& 0.863\scriptsize{$\pm$0.008}& 0.385\scriptsize{$\pm$0.099}& 0.260\scriptsize{$\pm$0.081}& 0.604\scriptsize{$\pm$0.044}& 0.592\scriptsize{$\pm$0.087}& 0.672\scriptsize{$\pm$0.059} \\
&Spatial-ID & 0.905\scriptsize{$\pm$0.009}& 0.225\scriptsize{$\pm$0.164}& 0.740\scriptsize{$\pm$0.029}& 0.437\scriptsize{$\pm$0.068}& 0.569\scriptsize{$\pm$0.063}& 0.859\scriptsize{$\pm$0.013}& 0.385\scriptsize{$\pm$0.080}& 0.227\scriptsize{$\pm$0.069}& 0.590\scriptsize{$\pm$0.030}& 0.430\scriptsize{$\pm$0.430}& 0.611\scriptsize{$\pm$0.055} \\
&STAGE&  0.903\scriptsize{$\pm$0.006}& 0.197\scriptsize{$\pm$0.127}& 0.738\scriptsize{$\pm$0.046}& 0.499\scriptsize{$\pm$0.103}& 0.535\scriptsize{$\pm$0.067}& 0.863\scriptsize{$\pm$0.008}& 0.385\scriptsize{$\pm$0.099}& 0.330\scriptsize{$\pm$0.172}& 0.597\scriptsize{$\pm$0.039}& 0.437\scriptsize{$\pm$0.034}& 0.674\scriptsize{$\pm$0.037}\\
&STANDS&  0.898\scriptsize{$\pm$0.005}& 0.111\scriptsize{$\pm$0.043}& 0.718\scriptsize{$\pm$0.019}& 0.453\scriptsize{$\pm$0.031}& 0.566\scriptsize{$\pm$0.015}& 0.863\scriptsize{$\pm$0.003}& 0.341\scriptsize{$\pm$0.050}& 0.079\scriptsize{$\pm$0.035}& 0.699\scriptsize{$\pm$0.059}& 0.604\scriptsize{$\pm$0.046}& 0.582\scriptsize{$\pm$0.049}\\
&TESLA& 0.820\scriptsize{$\pm$0.015}& 0.476\scriptsize{$\pm$0.020}& 0.404\scriptsize{$\pm$0.005}& 0.579\scriptsize{$\pm$0.022}& 0.643\scriptsize{$\pm$0.004}& 0.663\scriptsize{$\pm$0.018}& \underline{0.676}\scriptsize{$\pm$0.016}& 0.498\scriptsize{$\pm$0.008}& 0.641\scriptsize{$\pm$0.034}& 0.586\scriptsize{$\pm$0.016}& 0.510\scriptsize{$\pm$0.026}\\
&iStar& 0.803\scriptsize{$\pm$0.028}& 0.483\scriptsize{$\pm$0.026}& 0.389\scriptsize{$\pm$0.026}& 0.536\scriptsize{$\pm$0.044}& \underline{0.701}\scriptsize{$\pm$0.006}& 0.700\scriptsize{$\pm$0.027}& 0.618\scriptsize{$\pm$0.023}& 0.582\scriptsize{$\pm$0.009}& 0.668\scriptsize{$\pm$0.065}& \underline{0.689}\scriptsize{$\pm$0.020}& 0.135\scriptsize{$\pm$0.033}\\
&MEATRD& 0.907\scriptsize{$\pm$0.022}& \underline{0.886}\scriptsize{$\pm$0.021}& \textbf{0.843}\scriptsize{$\pm$0.007}& \underline{0.752}\scriptsize{$\pm$0.008}& 0.536\scriptsize{$\pm$0.013}& 0.850\scriptsize{$\pm$0.005}& 0.648\scriptsize{$\pm$0.017}& 0.453\scriptsize{$\pm$0.027}& 0.630\scriptsize{$\pm$0.001}& 0.333\scriptsize{$\pm$0.007}& 0.597\scriptsize{$\pm$0.003}\\
&SpaCell-Plus& \underline{0.956}\scriptsize{$\pm$0.003}& 0.778\scriptsize{$\pm$0.044}& \underline{0.827}\scriptsize{$\pm$0.016}& 0.730\scriptsize{$\pm$0.013}& 0.672\scriptsize{$\pm$0.042}& \underline{0.902}\scriptsize{$\pm$0.010}& 0.668\scriptsize{$\pm$0.043}& \underline{0.645}\scriptsize{$\pm$0.022}& \underline{0.766}\scriptsize{$\pm$0.025}& 0.667\scriptsize{$\pm$0.036}& \underline{0.758}\scriptsize{$\pm$0.009}\\
&SpaCRD(ours)& \textbf{0.975}\scriptsize{$\pm$0.002}& \textbf{0.926}\scriptsize{$\pm$0.012}& 0.818\scriptsize{$\pm$0.004}& \textbf{0.878}\scriptsize{$\pm$0.009}& \textbf{0.866}\scriptsize{$\pm$0.011}& \textbf{0.929}\scriptsize{$\pm$0.005}& \textbf{0.789}\scriptsize{$\pm$0.007}& \textbf{0.860}\scriptsize{$\pm$0.009}& \textbf{0.867}\scriptsize{$\pm$0.014}& \textbf{0.795}\scriptsize{$\pm$0.013}& \textbf{0.854}\scriptsize{$\pm$0.008}\\
\bottomrule
\end{tabular}}
\caption{Quantitative evaluation of SpaCRD for CTR detection on eleven breast cancer datasets compared to baselines. Each reported values are means ± standard deviations over five independent runs. Best results in bold, second-best underlined.}
\label{table2}
\end{table*}

\section{Methods \footnote{Related Work is in Supplementary Material S1 due to space limitation.}}
As shown in Figure~\ref{fig2}, our proposed SpaCRD framework consists of three main training stages:
(1) \textbf{Modality-alignment representation learning.} We employ a pre-trained pathology foundation model UNI \cite{uni} to extract informative histology image features. Meanwhile, we adopt a CLIP-based contrastive learning strategy \cite{clip} to align histology and ST modalities in a shared embedding space, effectively narrowing the modality gap and preparing for subsequent fusion.
(2) \textbf{VRBCA Fusion Network.} VRBCA aims to learn compact and informative representations of the interactions between histology and ST modalities.
(3) \textbf{Cancer likelihood estimation.} SpaCRD estimates a cancer likelihood score for each spot based on the compact representation yielded by VRBCA.
The overall framework is summarized in the algorithm in Supplementary Material S2.

\subsection{Modality-Alignment Representation Learning}
First, we crop image patches from histology images based on spatial coordinates of each spot in ST data. Patch size is determined by spot diameter in ST data and the pixel resolution of the histology images \cite{jaume2024transcriptomics}. We adopt UNI, a pathology-specific foundation model pretrained on large-scale histology data, as our histological feature extractor. Due to its demonstrated ability to capture fine-grained histological structures, we skip the fine-tuning step to reduce computational overhead. Let $I = \left\{ I_i \mid I_i \in \mathbb{R}^{l \times l \times 3} \right\}_{i=1}^{n}$ denote the set of image patches corresponding to all spots, where $n$ is the total number of spots, and $l=d/r$, with $d$ and $r$ representing the spot diameter and the pixel resolution of the histology image, respectively. Then passing the entire set of patches $I$ through the UNI model yields the H\&E embeddings $X^{\text{img}} = \left \{ \mathbf{x}_{1}^{\text{img}},\dots ,\mathbf{x}_{n}^{\text{img}} \right \}$ where 
\begin{equation}
\mathbf{x}_{i}^{\text{img}}=f_{\text{UNI}}\left ( I_{i} \right ) , \quad i=1,\dots ,n.
\end{equation}

\noindent Let $X^{\text{gene}}=\left \{ \mathbf{x}_{1}^{\text{gene}},\dots ,\mathbf{x}_{n}^{\text{gene}} \right \}$ represent the set of gene expression profiles in the ST data, where $\mathbf{x}_{i}^{\text{gene}}$ denotes the normalized expression vector of spot $i$. Given the substantial heterogeneity between image embeddings and gene expression data \cite{jaume2024modeling}, we employ a contrastive learning strategy to align the two modalities. By optimizing the contrastive loss, the model encourages paired image features and gene expression vectors from the same spatial location to be close in the latent space, while pushing apart those from different locations. This alignment reduces inconsistencies during the subsequent fusion process, ensuring a more stable and effective integration. We design two lightweight three-layer MLP-based encoders—an image encoder $f_{c1}$ and a gene encoder $f_{c2}$—to perform the aforementioned alignment task:
\begin{equation}
\mathbf{h}_i^{\text{img}} = f_{c1}(\mathbf{x}_i^{\text{img}}), \quad
\mathbf{h}_i^{\text{gene}} = f_{c2}(\mathbf{x}_i^{\text{gene}}).
\end{equation}
To measure similarity between latent representations, we construct a similarity matrix using their scaled dot-product:
\begin{equation}
\mathbf{S}_{ij} = \frac{ \mathbf{h}_i^{\text{img}} \cdot \mathbf{h}_j^{\text{gene}} }{\tau}, \quad
\text{where } \|\mathbf{h}_i^{\text{img}}\| = \|\mathbf{h}_j^{\text{gene}}\| = 1,
\end{equation}
with $\tau$ being a temperature parameter. Here, $\mathbf{S}_{ii}$ denotes the similarity between the $i$-th image and its paired gene expression, while $\mathbf{S}_{ij}$ for $i \ne j$ represents the similarity between unmatched pairs. Two directional InfoNCE loss components, $\mathcal{L}_{\text{img} \rightarrow \text{gene}}$ and $\mathcal{L}_{\text{gene} \rightarrow \text{img}}$, are independently computed using cross-entropy over similarity logits:
\begin{equation}
\mathcal{L}_{\text{img} \rightarrow \text{gene}} = -\frac{1}{n} \sum_{i=1}^{n} \log \frac{\exp(\mathbf{S}_{ii})}{\sum_{j=1}^{n} \exp(\mathbf{S}_{ij})},
\end{equation}
\begin{equation}
\mathcal{L}_{\text{gene} \rightarrow \text{img}} = -\frac{1}{n} \sum_{i=1}^{n} \log \frac{\exp(\mathbf{S}_{ii})}{\sum_{j=1}^{n} \exp(\mathbf{S}_{ji})}. 
\end{equation}
Finally, the total contrastive loss for this stage is defined as:
\begin{equation}
\mathcal{L}_{\text{contrast}} = \alpha \times  \mathcal{L}_{\text{img} \rightarrow \text{gene}} + \left ( 1-\alpha  \right )\times \mathcal{L}_{\text{gene} \rightarrow \text{img}},
\end{equation}
where $\alpha \in [0, 1]$ controls the balance between the two directional losses. For all experiments, $\alpha$ is fixed at 0.5.
\subsection{VRBCA Fusion Network}
Following modality-alignment representation learning, the encoded features are fed into the VRBCA network. Overall, VRBCA utilizes a bidirectional cross-attention (BCA) mechanism to comprehensively model interactions between spots and to integrate the aligned features from multiple perspectives, followed by a category-regularized variational autoencoder (RVAE) for filtering out noise and promoting the generation of compact and class-specific embeddings. We define two independent Cross-Attention (CA) modules with identical architectures: gene-guided and H\&E-guided CA blocks. Each module employs $m$ parallel attention heads. For the $i$-th head, the query, key, and value matrices are derived through linear projections of the inputs:
\begin{equation}
Q_i = Q W_i^Q, K_i = K W_i^K, V_i = V W_i^V.
\end{equation}
Here, the shapes of $W_i^Q$, $W_i^K$, and $W_i^V$ depend on the input modality and output embedding dimension $d$. The output of each head is computed as:
\begin{equation}
h_i = \text{Attention}(Q_i, K_i, V_i) = \text{softmax} \left( \frac{Q_i K_i^T}{\sqrt{d_h}} \right) V_i,
\end{equation}
\begin{equation}
\text{CA}(Q, K, V) = \left (  h_1 \mathbin\Vert h_2 \mathbin\Vert \cdots \mathbin\Vert h_m \right )  W,
\end{equation}
where $\mathbin\Vert$ denotes the concatenation of outputs from all attention heads, and $W$ is a shared projection matrix.

\begin{figure}[t]
\centering
\includegraphics[width=0.95\columnwidth]{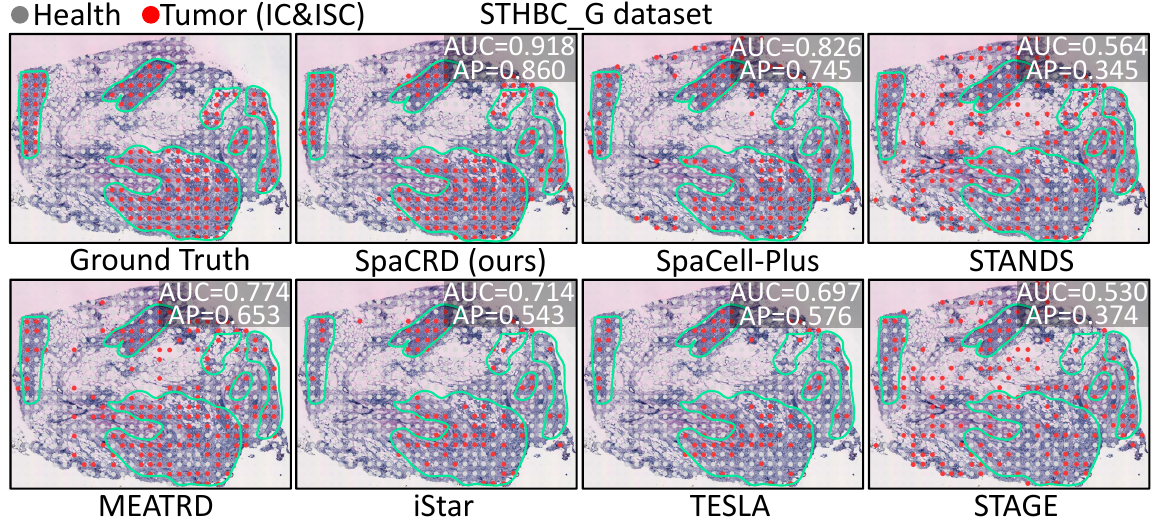}
\caption{ATR detection results of SpaCRD and other baselines on the STHBC\_G dataset. Green outlines indicate pathologist-annotated CTR. Gray and red dots represent normal and cancerous spots, respectively.}
\label{fig3}
\end{figure}

Next, let $H^{\text{img}}_{i}=\left [ \mathbf{h}_{i}^{\text{img}}, \mathbf{h}_{i_{1}}^{\text{img}},\dots , \mathbf{h}_{i_{k}}^{\text{img}} \right ] ^{T}$ and $H^{\text{gene}}_{i}=\left [ \mathbf{h}_{i}^{\text{gene}}, \mathbf{h}_{i_{1}}^{\text{gene}},\dots , \mathbf{h}_{i_{k}}^{\text{gene}} \right ] ^{T}$ denote the input matrices for spot $i$, where $\mathbf{h}_{i}^{\text{img}}$ and $\mathbf{h}_{i}^{\text{gene}}$ denote the contrastively aligned embeddings of H\&E image and gene expression for spot $i$, respectively, and $\left \{ i_1,\dots ,i_k \right \} \subset \mathcal{N}(i)$ denotes its set of neighboring spots. The operation of the BCA module proceeds:
\begin{equation}
Z_i^{\text{img}} = \text{CA}_{\text{img}}(H^{\text{img}}_{i}, H^{\text{gene}}_{i}, H^{\text{gene}}_{i}),
\end{equation}
\begin{equation}
Z_i^{\text{gene}} = \text{CA}_{\text{gene}}(H^{\text{gene}}_{i}, H^{\text{img}}_{i}, H^{\text{img}}_{i}),
\end{equation}
\begin{equation}
\mathbf{h}_i^{*} = \text{MLP}(Z_i^{\text{img}}[0] \mathbin\Vert Z_i^{\text{gene}}[0]),
\end{equation}
where $h_i^{*}$ denotes the multimodal representation of spot $i$, fused by the BCA module based on the features aligned through contrastive learning.

To further refine the fused multimodal representation $h_i^{*}$, the training of VRBCA network is guided by a category-regularized variational reconstruction objective, which imposes semantic structure on the latent space and facilitates the learning of class-discriminative multimodal features.
Specifically, we extend the VAE framework by introducing learnable class-specific latent centers, forming a structure-aware latent prior.
VRBCA encodes the fused multimodal representation into latent variables via an encoder $f_{\text{enc}}$, producing mean $\boldsymbol{\mu}_i$ and log-variance log $\boldsymbol{\sigma}_i^{2}$. A latent vector $\mathbf{z}_i\sim \mathcal{N}(\boldsymbol{\mu}_i, \boldsymbol{\sigma}_i^2)$ is sampled and decoded by $f_{\text{dec}}$ to reconstruct the multimodal representation, yielding $\hat{h}_i^{*}$:
\begin{equation}
\boldsymbol{\mu}_i, \text{log}\boldsymbol{\sigma}_i^{2}=f_{\text{enc}}\left ( \mathbf{h}_i^{*} \right ) ,
\end{equation}
\begin{equation}
z_{i}=\boldsymbol{\mu}_i + \boldsymbol{\sigma} \odot \epsilon , \epsilon \sim \mathcal{N}(\mathbf{0}, \mathbf{I}),
\end{equation}
\begin{equation}
\hat{\mathbf{h}}_i^{*} = f_{\text{dec}} \left ( z_{i} \right ) .
\end{equation}
The training objective for the fused representation is defined as:
\begin{equation}
\mathcal{L}_{\text{fused}} = \frac{1}{n} \sum_{i=1}^{n} \left [ \left\| \hat{\mathbf{h}}_i^{*} - \mathbf{h}_i^{*} \right\|^2 + \beta \cdot \mathcal{D}_{\text{KL}}^{\text{cls}} \left( q_i \,\|\, p_{y_i} \right) \right ] ,
\end{equation}
\begin{equation}
\scalebox{0.85}{$
\mathcal{D}_{\text{KL}}^{\text{cls}} (q_i \| p_{y_i}) = \tfrac{1}{2} \sum_j \left[ \sigma_{i,j}^2 + (\mu_{i,j} - \mu_{y_i,j})^2 - \log \sigma_{i,j}^2 - 1 \right],
$}
\end{equation}
where $q_i = \mathcal{N}(\boldsymbol{\mu}_i, \operatorname{diag}(\boldsymbol{\sigma}_i^2))$ is the approximate posterior, $p_{y_i} = \mathcal{N}(\boldsymbol{\mu}_{y_i}, \mathbf{I})$ is the class-specific Gaussian prior with learnable mean $\boldsymbol{\mu}_{y_i}$ based on the label $y_i \in \{0,1\}$. In all experiments, we set $\beta = 0.5$.

\subsection{Cancer Likelihood Discriminator}
After the modality-aware fusion by the VRBCA network, the cancer likelihood discriminator estimates the probability of each spot being cancerous. Specifically, the mean $\boldsymbol{\mu}_i$ and log-variance $\log \boldsymbol{\sigma}^2$ produced by the trained encoder $f_{\text{enc}}$ of VRBCA,  are concatenated and fed into a two-layer MLP classifier to predict cancer likelihood. The loss function for the discriminator combines BCE loss with the fused loss:
\begin{equation}
\scalebox{0.8}{$
\mathcal{L}_{\text{BCE}} = -\frac{1}{n} \sum_{i=1}^{n} \left[ y_i \cdot \log \sigma(\hat{y}_i) + (1 - y_i) \cdot \log (1 - \sigma(\hat{y}_i)) \right],
$}
\end{equation}
\begin{equation}
\mathcal{L}_{\text{cls}} = \mathcal{L}_{\text{BCE}} + \gamma \times \mathcal{L}_{\text{fused}},
\end{equation}
where $\hat{y}_i$ denotes the predicted score by the classifier indicating how likely the spot is cancerous. The hyperparameters $\gamma$ is fixed to $0.1$ in the experiments. At inference time, benefiting from the regularized latent representations, our model effectively separates cancerous and normal spots in the predicted score space. To automatically determine a classification threshold, we fit a two-component Gaussian Mixture Model (GMM) to the distribution of predicted scores and use the intersection point between the two Gaussian components as the decision threshold, as detailed in Supplementary Material S3.

\begin{figure}[!t]
\centering
\includegraphics[width=0.75\columnwidth]{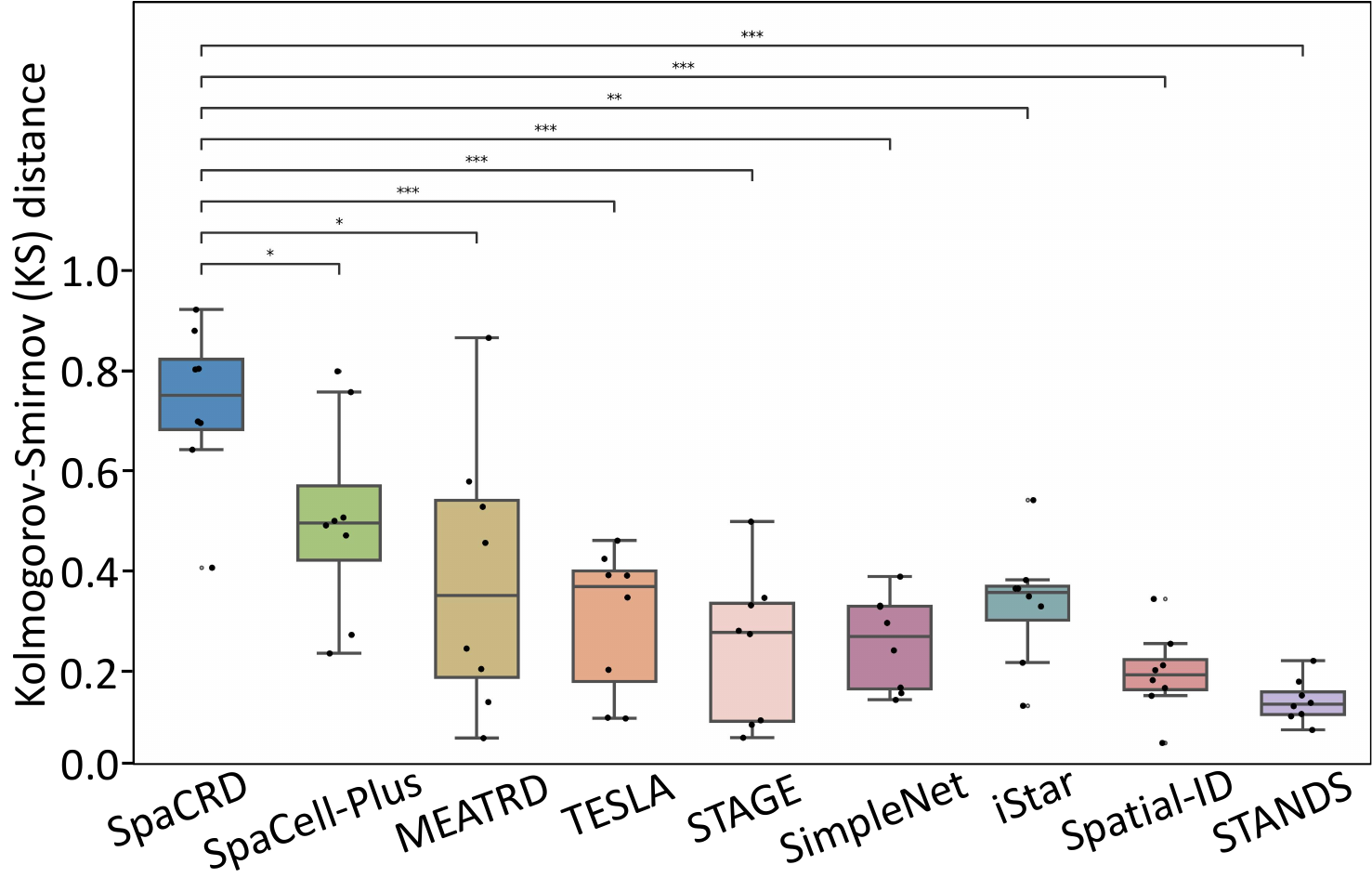}
\caption{Comparison of the KS distances between predicted cancer likelihood distributions in healthy and tumor regions.}
\label{ks}
\end{figure}

\begin{figure}[t]
\centering
\includegraphics[width=1\columnwidth]{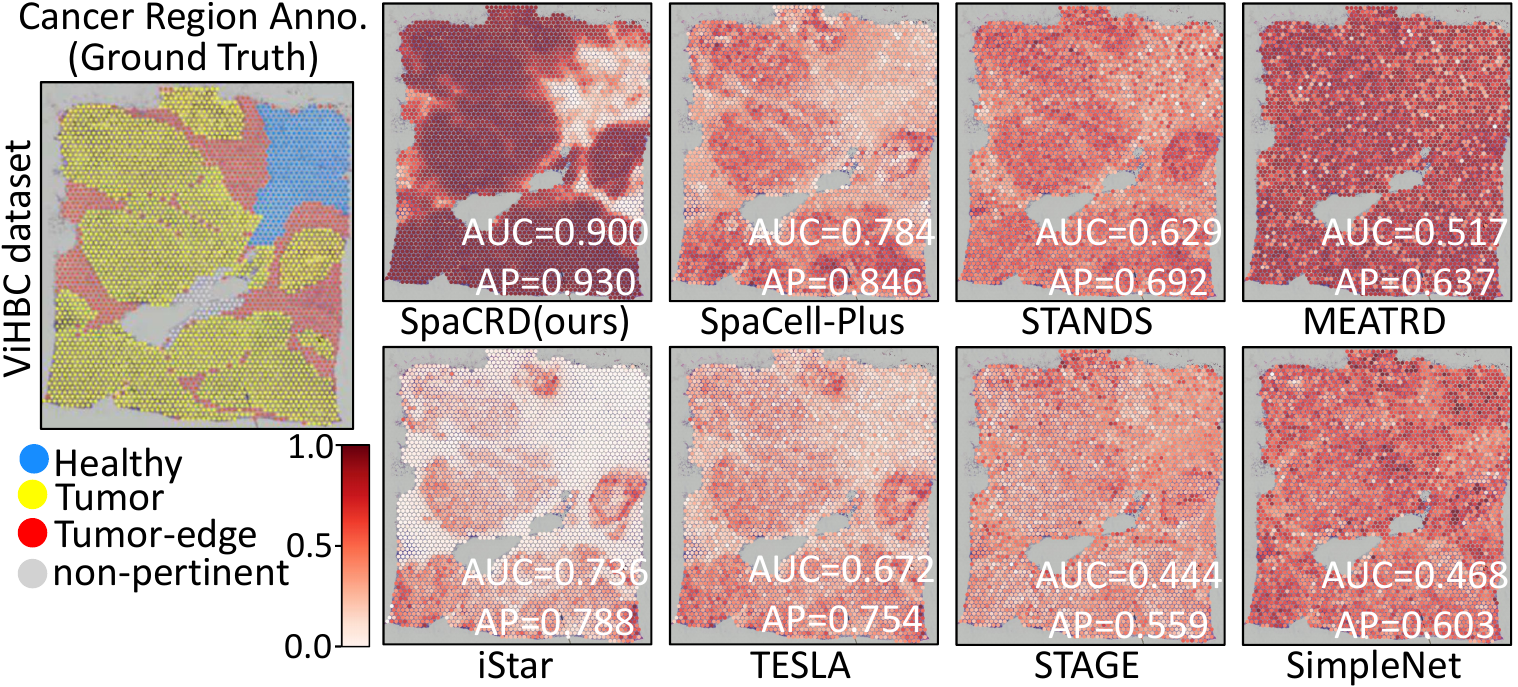}
\caption{Visualization of cancer likelihood scores predicted by SpaCRD and other baselines on the ViHBC dataset.}
\label{fig4}
\end{figure}

\section{Experiments}
\subsection{Experimental Setup}
\subsubsection{Datasets.}
We evaluate SpaCRD on five datasets comprising a total of 23 matched histology-ST datasets (termed tissue sections). Among them, two multi-section datasets-STHBC \cite{her2st} and CRC \cite{crc}-are used for cross-sample evaluation, while the remaining three—10XHBC, XeHBC \cite{xenium}, and IDC-are used for cross-platforms\&batches evaluation (see Supplementary Material S4 for detailed dataset description and preprocessing of the 23 tissue sections).

\subsubsection{Implementation Details.}
We conducted all experiments using a single NVIDIA RTX 3090 GPU (24GB), with the development environment based on PyTorch 2.1.1 and Python 3.11.5. Detailed network architectures, training schedules, and implementation settings are provided in Supplementary Material S5.

\subsubsection{Baselines and Evaluation Metrics.}
To evaluate the performance of SpaCRD, we compared it with eight SOTA methods, including five multimodal-based: MEATRD \cite{xu2025meatrd}, STANDS \cite{xu2024detecting}, SpaCell \cite{tan2020spacell}, iStar \cite{istar}, TESLA \cite{tesla}; two ST-based: STAGE \cite{stage}, Spatial-ID \cite{shen2022spatial}; and one image-based: SimpleNet \cite{liu2023simplenet}. For evaluation metrics, AUC, AP, F1-score, and KS distance are used to assess the performance of all methods. Supplementary Materials S6 and S7 provide the detailed descriptions for baselines and evaluation metrics.

\begin{figure}[!t]
\centering
\includegraphics[width=0.9\columnwidth]{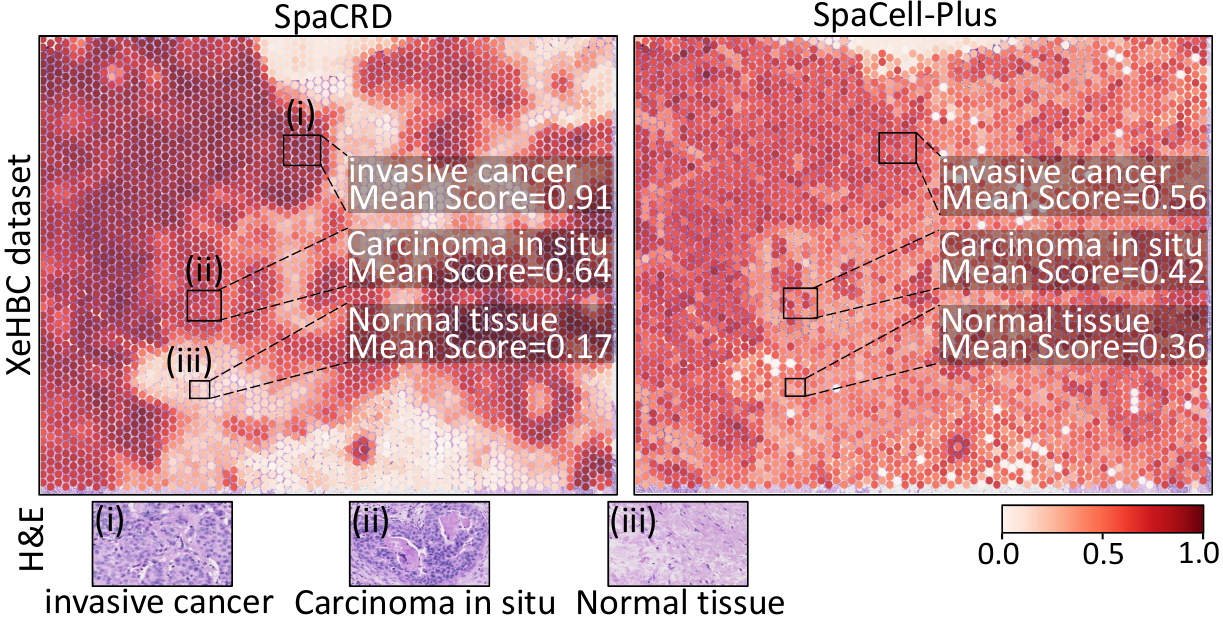}
\caption{Cancer likelihood scores predicted by SpaCRD and the best-performing baseline method, SpaCell-Plus on the XeHBC dataset.}
\label{fig5}
\end{figure}

\begin{figure}[t]
\centering
\includegraphics[width=0.9\columnwidth]{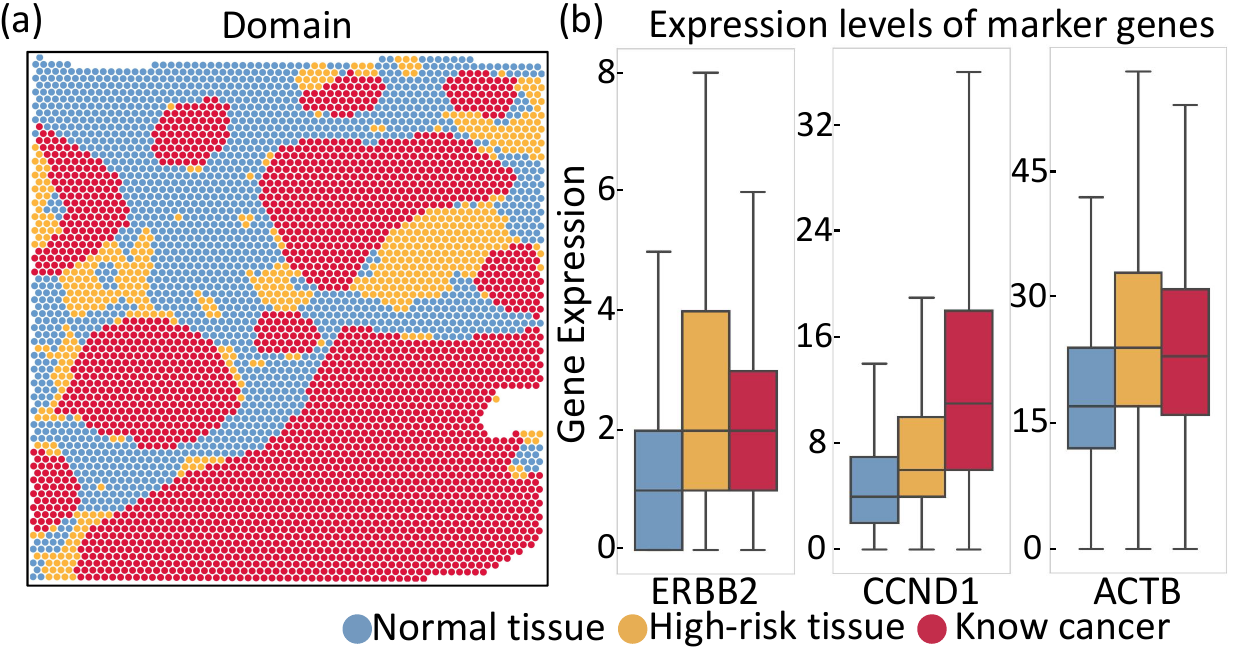}
\caption{In the IDC dataset, SpaCRD-predicted high-score spots labeled non-cancerous exhibit elevated breast cancer marker expression, suggesting potential early lesions.}
\label{fig6}
\end{figure}

\subsection{Cross-Samples Cancer Region Detection}
To evaluate the cross-samples CTR detection capability of SpaCRD, we conducted leave-one-out cross-validation on the twelve colorectal cancer (CRC\_[A1-G2]) and eight human breast cancer (STHBC\_[A-H]) datasets. As shown in Table~\ref{table1} and \ref{table2} (Cross-samples), SpaCRD achieves the best performance in AUC, AP, and F1-score, surpassing all baselines. Across 20 datasets covering both breast cancer and colorectal cancer, SpaCRD consistently outperforms the second-best method, achieving an average improvement of 13.5\%, 14.1\%, and 14.0\% in AUC, AP, and F1-score, respectively. Methods that rely on prior knowledge, such as iStar and TESLA, perform suboptimally, possibly due to limitations in generalizing predefined features across datasets. Similarly, methods like MEATRD and STANDS, which adopt training strategies inspired by conventional anomaly detection in computer vision, struggle to capture the continuous and structured nature of cancer regions. SpaCell-Plus lacks effective integration of histology and ST data, resulting in inferior performance compared with our method. We further demonstrate the superiority of SpaCRD over the baselines by visualizing the CTR detection results on the STHBC datasets (Figure~\ref{fig3} and Supplementary Material S8.1). Finally, we evaluated the distributional divergence between predicted cancer likelihood scores in healthy and tumor regions across the eight STHBC datasets using the Kolmogorov–Smirnov (KS) distance. As shown in Figure~\ref{ks}, SpaCRD exhibited the strongest separation between healthy and tumor regions across most STHBC datasets (Median: SpaCRD=0.754, SpaCell-Plus=0.494, MEATRD=0.348).

\subsection{Cross-Platforms\&batches Cancer Region Detection}
We further evaluated the performance of SpaCRD on the challenging task of cross-platforms\&batches CTR detection. We trained SpaCRD on the STHBC dataset generated from the ST platform and tested it on datasets from other platforms, including ViHBC (Visium), IDC (Visium) and XeHBC (Xenium) datasets. As shown in Table~\ref{table2} (Cross-platforms\&batches), SpaCRD achieves superior performance over all baselines across all test datasets, with average improvements of 12.1\%, 11.8\%, and 13.8\% in AUC, AP, and F1-score, respectively, compared to the second-best method. In addition, we comparative analyzed the predicted cancer likelihood scores of SpaCRD and other baseline methods on the ViHBC dataset. As shown in Figure~\ref{fig4}, SpaCRD accurately distinguishes cancer regions from healthy tissues. Moreover, compared to baselines, it effectively separates the tumor-edge in certain regions, with predicted scores falling between those of the cancer and healthy spots. SpaCell-Plus and STANDS also identified ATR regions; however, the separation between cancer and healthy regions was less distinct compared to SpaCRD. iStar and TESLA were able to predict parts of the cancer regions; however, their predictions were incomplete and failed to capture the full extent of the lesions. Distribution analysis using histograms and violin plots further confirms SpaCRD’s ability to distinguish between biologically distinct tissue regions (see Supplementary Material S8.2). Consistent with the findings on the ViHBC dataset, visualization analysis of the XeHBC and IDC datasets shows that the cancer likelihood scores predicted by SpaCRD are highly consistent with ground truth (Supplementary Material S8.1). These results suggest SpaCRD mitigates technical and batch effects by integrating image and ST data into a unified latent space, enabling robust, generalizable CTR detection across diverse platforms and batches.

\subsection{Downstream Analysis: Detection of Potential Lesion Regions}
\begin{table*}[!t]
\centering
\resizebox{0.8\textwidth}{!}{
\begin{tabular}{l|cccccc|ccccccccc}
\toprule
\multicolumn{1}{c|}{\multirow{3}{*}{Extractor}} & \multicolumn{6}{c|}{\textbf{Cross-samples}} & \multicolumn{9}{c}{\textbf{Cross-platforms\&batches}}\\ \cline{2-16} 
& \multicolumn{3}{c}{STHBC} & \multicolumn{3}{c|}{CRC} & \multicolumn{3}{c}{10XHBC} & \multicolumn{3}{c}{IDC} & \multicolumn{3}{c}{XeHBC}\\ \cline{2-16} 
& AUC& AP& F1& AUC& AP& F1& AUC& AP& F1& AUC& AP& F1& AUC& AP& F1\\ \hline
w/  Swin-Tiny & 0.880& 0.923& 0.838& 0.820& 0.795& 0.793& 0.689& 0.751& 0.728& 0.501& 0.601& 0.601 & 0.718& 0.465& 0.501\\
w/  ResNet50  & 0.878& 0.908& 0.836& 0.830& 0.800& 0.794& 0.637& 0.697& 0.701& 0.596& 0.649& 0.657 & 0.656& 0.698& 0.691\\
w/  HIPT      & 0.885& 0.912& 0.827& 0.822& 0.796& 0.793& 0.784& 0.829& 0.786& 0.399& 0.525& 0.546& 0.771& 0.736& 0.742\\
w/ UNI (ours) & \textbf{0.929}& \textbf{0.946}& \textbf{0.880}& \textbf{0.869}& \textbf{0.854}& \textbf{0.810}& \textbf{0.900}& \textbf{0.930}& \textbf{0.867}& \textbf{0.891}& \textbf{0.914}& \textbf{0.854} & \textbf{0.931}& \textbf{0.859}& \textbf{0.795}\\
\bottomrule
\end{tabular}}
\caption{Ablation study of histology feature extractors conducted across all datasets used in this study.}
\label{ablation_1}
\end{table*}

\begin{table}[t]
\centering
\resizebox{0.8\columnwidth}{!}{
\begin{tabular}{l|cccccc}
\toprule
\multicolumn{1}{c|}{\multirow{2}{*}{Model}} & \multicolumn{3}{c}{HBC(eleven datasets)}& \multicolumn{3}{c}{CRC(twelve datasets)}\\ \cline{2-7} 
\multicolumn{1}{c|}{}                         & \multicolumn{1}{c}{AUC} & \multicolumn{1}{c}{AP} & \multicolumn{1}{c}{F1} & \multicolumn{1}{c}{AUC} & \multicolumn{1}{c}{AP} & \multicolumn{1}{c}{F1} \\ \hline
Image-based   & 0.789& 0.752& 0.733& 0.606& 0.557& 0.553\\
ST-based      & 0.832& 0.815& 0.747& 0.782& 0.793& 0.755\\
w/o  BCA      & 0.849& 0.833& 0.788& 0.797& 0.774& 0.759\\
w/o  RVAE      & 0.887& 0.898& 0.817& 0.831& 0.816& 0.795\\
w/o  VRBCA      & 0.815& 0.796& 0.734& 0.771& 0.746& 0.717\\
w/o  CL       & 0.892& 0.886& 0.828& 0.824& 0.807& 0.776\\ \hline
\textbf{Ours} & \textbf{0.923}& \textbf{0.934}& \textbf{0.869}& \textbf{0.869}& \textbf{0.854}& \textbf{0.810}\\
\bottomrule
\end{tabular}}
\caption{Ablation study of modalities and fusion modules conducted across all datasets used in this study.}
\label{ablation_2}
\end{table}

\begin{table}[t]
\centering
\resizebox{0.75\columnwidth}{!}{
\begin{tabular}{ccccccc}
\toprule
\multicolumn{1}{c}{\multirow{2}{*}{Metric}} & \multicolumn{3}{c}{Parameter $\alpha$}& \multicolumn{3}{c}{Parameter $\beta $} \\ \cline{2-7} 
\multicolumn{1}{c}{}                         & \multicolumn{1}{c}{0.0} & \multicolumn{1}{c}{0.5} & \multicolumn{1}{c}{1.0} & \multicolumn{1}{c}{0.1} & \multicolumn{1}{c}{0.5} & \multicolumn{1}{c}{1.0} \\ \hline
AUC   & 0.908& \cellcolor{gray!30}\textbf{0.923}& 0.916& 0.863& \cellcolor{gray!30}\textbf{0.923}& 0.920 \\
AP   & 0.912& \cellcolor{gray!30}\textbf{0.934}& 0.929& 0.895& \cellcolor{gray!30}\textbf{0.934}& 0.931 \\
F1   & 0.843& \cellcolor{gray!30}\textbf{0.869}& 0.863& 0.828& \cellcolor{gray!30}\textbf{0.869}& 0.867 \\ \hline
\multicolumn{1}{c}{\multirow{2}{*}{Metric}} & \multicolumn{3}{c}{Parameter $\gamma $}& \multicolumn{3}{c}{neighboring spots}  \\ \cline{2-7} 
\multicolumn{1}{c}{}                         & \multicolumn{1}{c}{0.0} & \multicolumn{1}{c}{0.1} & \multicolumn{1}{c}{1.0} & \multicolumn{1}{c}{4} & \multicolumn{1}{c}{6} & \multicolumn{1}{c}{10}  \\ \hline
AUC   & 0.772& \cellcolor{gray!30}\textbf{0.923}& 0.919& 0.908& \cellcolor{gray!30}\textbf{0.923}& 0.912\\
AP   & 0.768& \cellcolor{gray!30}\textbf{0.934}& 0.931& 0.896& \cellcolor{gray!30}\textbf{0.934}& 0.925\\
F1   & 0.794& \cellcolor{gray!30}\textbf{0.869}& 0.856& 0.844& \cellcolor{gray!30}\textbf{0.869}& 0.858\\
\bottomrule
\end{tabular}}
\caption{Sensitivity analysis of key hyperparameter across eleven breast cancer datasets. \colorbox{gray!30}{Gray}: default settings.}
\label{Sensitivity}
\end{table}
We systematically evaluated SpaCRD's ability to stratify cancer severity by analyzing its predicted likelihood scores. On the XeHBC dataset generated from Xenium platform, SpaCRD achieved average scores of 0.91 (invasive cancer), 0.64 (carcinoma in situ), and 0.17 (normal tissue), effectively distinguishing malignancy levels (Figure~\ref{fig5}). Heatmap visualization confirmed clear stratification, with invasive regions showing the highest intensity, followed by carcinoma in situ and normal tissue. In contrast, SpaCell-Plus produced less discriminative scores (0.56, 0.42, and 0.36, respectively), failing to separate carcinoma in situ from normal tissue. Other baselines also exhibited dispersed, inconsistent score distributions (Supplementary Material S8.1). Notably, these distinctions are often imperceptible in histology images, suggesting SpaCRD's scores may reflect tumor aggressiveness
. Finally, we examined the spots in the IDC dataset that received high predicted scores from SpaCRD but were annotated as non-cancerous in the manual labels (Figure~\ref{fig6}(a), orange spots). Remarkably, these spots exhibit significantly elevated expression of canonical breast cancer marker genes (e.g., ERBB2 \cite{erbb2}, CCND1 \cite{CCND1}, and ACTB \cite{actb}, Figure~\ref{fig6}(b)), compared to normal tissue. This suggests that SpaCRD captures spatial regions warranting clinical investigation and potentially of pathological relevance.

\subsection{Ablation Studies}
To assess the contributions of key components in SpaCRD, we conducted ablation studies on all datasets, focusing on the Histology feature extractor, unimodal inputs, and module ablations. All reported results are averaged over five independent runs. \textbf{Since some experiments involve aggregating scores across multiple datasets, we report the overall mean values without standard deviations for consistency.}

\subsubsection{Impact of Extractors.}
We replaced UNI with ResNet50 \cite{resnet} and Swin-Tiny \cite{swin} pretrained on natural images, and HIPT \cite{hipt} pretrained on pathological images. As shown in Table~\ref{ablation_1}, UNI yields the best performance. Notably, the performance of other extractors drops markedly in the cross-platforms\&batches evaluation, likely due to large variations in histology images that hinder their ability to capture core fine-grained features.

\subsubsection{Impact of Modalities and Fusion Modules.} We assessed the contributions of different input modalities and the fusion module by (i) using only histology or ST data, and (ii) removing key fusion components, including the multimodal deep fusion module (BCA), the regularized reconstruction-guided denoising module (RVAE), and the modality-alignment representation learning (CL). As shown in Table~\ref{ablation_2}, removing any modality input or module results in suboptimal performance.

\subsection{Robustness and Efficiency Analysis}
\begin{itemize}
\item \textbf{Sensitivity Analysis}: We conducted sensitivity analyses on eleven breast cancer datasets, covering key hyperparameter $\alpha$, $\beta $, and $\gamma $, as well as the number of neighboring spots selected in the BCA model (Table~\ref{Sensitivity}). Detailed analyses are provided in Supplementary Material S8.3. 
\item \textbf{Empirical Efficiency Analysis}: We analyzed the cost of SpaCRD and baselines by evaluating the number of parameters, runtime, and memory usage. The detailed results in Supplementary Material S8.4 indicate that SpaCRD maintains an acceptable computational cost. 
\item \textbf{Small-Sample Training Analysis}: Additional experiments show that even when training set contains less than 10\% of the spots in test set, SpaCRD maintains stable performance in cross-platform CTR detection, demonstrating strong generalization and data efficiency (see Supplementary Material S8.5 for detailed results). 
\end{itemize}

\section{Conclusion}
In this study, we present SpaCRD, a multimodal deep fusion and transfer learning-based framework that integrates histology images and ST data for accurate and generalizable CTR detection across samples and platforms\&batches. SpaCRD leverages the proposed VRBCA fusion module, synergistically optimized through contrastive learning objectives, to dynamically integrate histology images and gene expression features in both image-to-gene and gene-to-image directions. By jointly attending to both gene-to-image and image-to-gene signals and modeling interactions among neighboring spots, VRBCA ensures comprehensive feature integration, while its variational reconstruction mechanism filters out noise and promotes compact and class-consistent embeddings. Extensive evaluations across diverse datasets from multiple platforms demonstrate the strong generalization capability of SpaCRD. In addition, SpaCRD may serve as a promising tool for medical researchers by enabling the stratification of cancer severity and revealing spatial regions that may warrant further clinical investigation.
Supplementary Materials can be found at the code link.

\section*{Acknowledgments}
The work was supported in part by the Program of Yunnan Key Laboratory of Intelligent Systems and Computing (No. 202405AV340009) and the National Natural Science Foundation of China (Nos. 62262069, 62571555).

\bibliography{aaai2026}
\clearpage

\setcounter{section}{0}
\setcounter{figure}{0}
\setcounter{table}{0}
\setcounter{equation}{0}
\renewcommand{\thesection}{S\arabic{section}}
\renewcommand{\thefigure}{S\arabic{figure}}
\renewcommand{\thetable}{S\arabic{table}}
\renewcommand{\theequation}{S\arabic{equation}}

\begin{center}
    \Large \textbf{Supplementary Material}
\end{center}

\section{Related Works}
\subsection{Cancer Tissue Region Detection}
Early CTR detection relied on traditional anomaly detection in histology images. For example, SimpleNet \cite{liu2023simplenet} generates pseudo-anomalies by perturbing visual embeddings with noise and employs a hyperplane-based discriminator for anomaly separation. However, such methods heavily rely on effective representation learning \cite{sohn2020learning}, struggling to handle batch effects commonly present in pathological images.

Spatial-ID \cite{shen2022spatial} is the first method to perform CTR detection using spatial transcriptomics (ST) data alone, which integrates transfer learning with a deep neural network trained on scRNA-seq data and employs a graph convolutional network \cite{kipf2016semi} to incorporate spatial context for improved cell typing and anomaly identification. However, due to its exclusive reliance on ST data, Spatial-ID is vulnerable to noise and dropout, which may lead to the misclassification of normal regions as abnormal when subtle expression similarities exist among healthy.

Super-resolution annotation methods that rely on expert-defined priors, such as iStar \cite{istar}, leverage pretrained HIPT \cite{hipt} models to extract hierarchical histological features, followed by a feedforward neural network to model histology and gene expression jointly, and then use an expert-defined list of marker genes to infer CTR. TESLA \cite{tesla} adopts a similar CTR detection strategy to iStar, but generates super-resolved gene expression by leveraging a Euclidean metric that integrates spatial proximity and histological similarity. The heavy dependence on expert-defined priors significantly constrains the wide applicability of these approaches.

SpaCell \cite{tan2020spacell} designs two independent autoencoders \cite{kingma2013auto} to integrate histology images and ST data into a unified latent representation for CTR detection. However, it lacks deep multimodal interaction and fails to generalize to cross-platform and cross-batch CTR detection scenarios. STANDS \cite{xu2024detecting} encodes histology and gene expression data using a hybrid network of graph attention (GAT) \cite{velickovic2017graph} and ResNet-GAT, followed by a Transformer-based fusion module to identify CTR. Following the training strategy of STANDS, MEATRD \cite{xu2025meatrd} combines Mobile-UNet, graph neural networks, and Transformer modules \cite{vaswani2017attention} in a unified architecture, and follows the conventional visual anomaly detection strategy by detecting CTR through reconstruction errors. However, due to the continuous and structured nature of cancer regions, reconstruction-error-based STANDS and MEATRD often fail to effectively distinguish them from normal tissue.

\subsection{Multimodal Integration of Histology and Spatial Transcriptomics}
Beyond CTR detection, multimodal integration of histology images and ST data has also been developed for related tasks, such as gene expression prediction \cite{min2024multimodal}, and spatial domain identification \cite{hu2024benchmarking,spagcn}.

mclSTExp \cite{min2024multimodal} predicts spatial gene expression from H\&E-stained whole-slide images by aligning histology and transcriptomics via contrastive learning. However, it performs only modality alignment without learning joint representations, effectively mapping one modality to the other rather than achieving true fusion.

SpaGCN \cite{spagcn} integrates gene expression and histology via a 3D weighted graph and applies Louvain clustering on GCN features to identify spatial domains. However, as these methods typically prioritize maximizing spatial variability, they are not well suited for CTR detection, which demands higher accuracy and interpretability.

In summary, existing multimodal approaches focus primarily on tasks like gene expression prediction or spatial domain identification, often employing alignment or simple aggregation techniques. While pioneering, these methods do not explicitly address the high-stakes demands of high-accuracy CTR detection, which requires robust fusion and precise, interpretable identification of clinically relevant boundaries, moving beyond maximizing general spatial variance.

\section{Algorithm Framework of SpaCRD}
The overall training and inference pipeline of SpaCRD is summarized in Algorithm \ref{alg:SpaCRD}.

\section{Determining Cancer Likelihood Score Threshold}  

To convert the predicted continuous scores (ranging from 0 to 1) into binary labels indicating cancerous versus healthy regions, we adopt a thresholding approach based on a Gaussian Mixture Model (GMM). This method is particularly suitable when the prediction score distribution is bimodal and manual thresholding proves unreliable.

We assume that the distribution of the model output scores $\hat{y}$ can be approximated by a mixture of two univariate Gaussian distributions, corresponding to the underlying healthy and cancerous tissue classes, respectively. The probability density function is given by:
\begin{equation}
p(\hat{y}) = \pi_1 \cdot \mathcal{N}(\hat{y} \mid \mu_1, \sigma_1^2) + \pi_2 \cdot \mathcal{N}(\hat{y} \mid \mu_2, \sigma_2^2),
\end{equation}
where $\mathcal{N}(\hat{y} \mid \mu, \sigma^2)$ denotes the Gaussian (normal) probability density function:
\begin{equation}
\mathcal{N}(\hat{y} \mid \mu, \sigma^2) = \frac{1}{\sqrt{2\pi\sigma^2}} \exp\left(-\frac{(\hat{y} - \mu)^2}{2\sigma^2}\right).
\end{equation}
In the mixture model, $\pi_1, \pi_2 \in (0,1)$ are the mixing coefficients (with $\pi_1 + \pi_2 = 1$), and $(\mu_k, \sigma_k^2)$ are the mean and variance of the $k$-th component ($k=1,2$).

The parameters $\Theta = \{ \pi_k, \mu_k, \sigma_k^2 \}_{k=1}^2$ are estimated from the data using the Expectation-Maximization (EM) algorithm, which maximizes the log-likelihood of the observed scores $\{\hat{y}_i\}_{i=1}^n$:

\begin{algorithm}[hbp]
\caption{SpaCRD: Spatial Cancer Region Detection Framework}
\label{alg:SpaCRD}
\textbf{Input}: Histology patches $I = \left\{ I_i \mid I_i \in \mathbb{R}^{l \times l \times 3} \right\}_{i=1}^{n}$; gene expression $X^{\text{gene}}=\left \{ X_{1}^{\text{gene}},\dots ,X_{n}^{\text{gene}} \right \}$, spot coordinates $S=\{\mathbf{s}_i\}_{i=1}^{n}$\\

\textbf{Definition}: Pre-trained pathology-specific foundation model $f_{\text{UNI}}$; Image encoder $f_{c1}$; Gene encoder $f_{c2}$; Two directional InfoNCE loss $\mathcal{L}_{\text{img} \rightarrow \text{gene}}$ and $\mathcal{L}_{\text{gene} \rightarrow \text{img}}$; Gene-guided cross-attention network $\text{CA}_{\text{gene}}$; H\&E-guided cross-attention network $\text{CA}_{\text{img}}$; Feedforward fusion network $f$; RVAE encoder $f_{\text{enc}}$ and decoder $f_{\text{dec}}$; L2 loss function $\mathcal{L}_{2}$; category-regularized KL divergence loss function $\mathcal{D}_{\text{KL}}^{\text{cls}}$; Cross-entropy loss function $\mathcal{L}_{\text{BCE}}$; MLP-based classifier $f_{\text{cls}}$.\\

\textbf{Output}: Cancer likelihood scores $\hat{\mathbf{y}}=\{\hat{y}_i\}_{i=1}^n$, binary labels $\hat{\mathbf{c}}=\{\hat{c}_i\}_{i=1}^n$
\begin{algorithmic}[1]

\FOR{each batch $b$}
    \STATE $X_b^{\text{img}} = \left\{ f_{\text{UNI}}(I_i) \mid i \in \mathcal{B}_b \right\}$
\ENDFOR
\FOR{each batch $b$}
    \STATE $H_b^{\text{img}} \gets f_{c1}(X_b^{\text{img}})$, $H_b^{\text{gene}} \gets f_{c2}(X_b^{\text{gene}})$
    \STATE $\mathcal{L}_{\text{contrast}} = \alpha \times  \mathcal{L}_{\text{img} \rightarrow \text{gene}} + \left ( 1-\alpha  \right )\times \mathcal{L}_{\text{gene} \rightarrow \text{img}}$
    \STATE Update parameters of $f_{c1}$, $f_{c2}$ using $\mathcal{L}_{\text{contrast}}$
\ENDFOR

\FOR{each batch $b$}
    \STATE $H_{bi}^{\text{img}} \gets$ image features of spot $i$ and $s$ neighbors
    \STATE $H_{bi}^{\text{gene}} \gets$ gene features of spot $i$ and $s$ neighbors
    \STATE $H_b^{m} = \left\{ H_{bi}^{m} \mid i \in \mathcal{B}_b,\ H_{bi}^{m} \in \mathbb{R}^{(s+1) \times d} \right\},\quad m \in \{\text{img}, \text{gene}\}$
    \STATE $Z_b^{\text{img}} = \text{CA}_{\text{img}}(H_b^{\text{img}}, H_b^{\text{gene}}, H_b^{\text{gene}})$
    \STATE $Z_b^{\text{gene}} = \text{CA}_{\text{gene}}(H_b^{\text{gene}}, H_b^{\text{img}}, H_b^{\text{img}})$
    \STATE $H_b^{*} = f(Z_b^{\text{img}}[:, 0, :] \mathbin\Vert Z_b^{\text{gene}}[:, 0, :])$
    \STATE $(\boldsymbol{\mu}_b, \log \boldsymbol{\sigma}_b^2) \gets f_{\text{enc}}(H_b^{*})$
    \STATE Sample latent $Z_b \sim \mathcal{N}(\boldsymbol{\mu}_b, \boldsymbol{\sigma}_b^2)$, $\hat{H}_b^* \gets f_{\text{dec}}(Z_b)$
    \STATE $\mathcal{L}_{\text{fused}} = \mathcal{L}_{2}(H_b^*, \hat{H}_b^*) + \beta \cdot \mathcal{D}_{\text{KL}}^{\text{cls}}(q_b \| p_y)$
    \STATE Update parameters of $\text{CA}_{\text{img}}$, $\text{CA}_{\text{gene}}$, $f$, $f_{\text{enc}}$, $f_{\text{dec}}$ using $\mathcal{L}_{\text{fused}}$
\ENDFOR

\FOR{each batch $b$}
    \STATE $\hat{\mathbf{y}}_b=\left\{ f_{\text{cls}}(\boldsymbol{\mu}_i \mathbin\Vert \log \boldsymbol{\sigma}_i^2) \mid i \in \mathcal{B}_b \right\}$
    \STATE $\mathcal{L}_{\text{cls}} = \mathcal{L}_{\text{BCE}} + \gamma \cdot \mathcal{L}_{\text{fused}}$
    \STATE Update parameters of $\text{CA}_{\text{img}}$, $\text{CA}_{\text{gene}}$, $f$, $f_{\text{enc}}$, $f_{\text{dec}}$, $f_{\text{cls}}$ using $\mathcal{L}_{\text{cls}}$
\ENDFOR
\STATE Fit GMM to $\hat{\mathbf{y}}$ and compute threshold $\tau$
\FOR{$i = 1$ to $n$}
    \STATE $\hat{c}_i \gets \mathbb{I}(\hat{y}_i \ge \tau)$
\ENDFOR

\STATE \textbf{return} $\hat{\mathbf{y}}$, $\hat{\mathbf{c}}$
\end{algorithmic}
\end{algorithm}

\begin{equation}
\begin{aligned}
\log p(\hat{y} \mid \Theta) =
& \sum_{i=1}^{n} \log \bigg[ \pi_1 \cdot \mathcal{N}(\hat{y}_i \mid \mu_1, \sigma_1^2) \\
& + \pi_2 \cdot \mathcal{N}(\hat{y}_i \mid \mu_2, \sigma_2^2) \bigg].
\end{aligned}
\end{equation}

By expanding both sides of the equality and simplifying, we obtain the nonlinear equation to solve for the decision threshold $\tau$, which corresponds to the intersection point(s) of the two Gaussian components:

\begin{equation}
\pi_1 \cdot \mathcal{N}(\hat{y} \mid \mu_1, \sigma_1^2) = \pi_2 \cdot \mathcal{N}(\hat{y} \mid \mu_2, \sigma_2^2).
\end{equation}
Expanding the Gaussian densities explicitly, this can be written as:
\begin{equation}
\frac{\pi_1}{\sqrt{2\pi\sigma_1^2}} \exp\left(-\frac{(\hat{y} - \mu_1)^2}{2\sigma_1^2} \right) = \frac{\pi_2}{\sqrt{2\pi\sigma_2^2}} \exp\left(-\frac{(\hat{y} - \mu_2)^2}{2\sigma_2^2} \right).
\end{equation}
Taking the natural logarithm of both sides yields:
\begin{equation}
\log \left( \frac{\pi_1}{\sigma_1} \right) - \frac{(\hat{y} - \mu_1)^2}{2\sigma_1^2}
=
\log \left( \frac{\pi_2}{\sigma_2} \right) - \frac{(\hat{y} - \mu_2)^2}{2\sigma_2^2}.
\end{equation}
Rearranging terms gives a quadratic equation in $\hat{y}$:
\begin{equation}
a \hat{y}^2 + b \hat{y} + c = 0,
\end{equation}
where the coefficients $(a,b,c)$ are determined analytically from $(\pi_k, \mu_k, \sigma_k^2)$:
\begin{equation}
\begin{gathered}
a = \frac{1}{2\sigma_2^2} - \frac{1}{2\sigma_1^2}, \\
b = \frac{\mu_2}{\sigma_2^2} - \frac{\mu_1}{\sigma_1^2}, \\
c = \frac{\mu_1^2}{2\sigma_1^2} - \frac{\mu_2^2}{2\sigma_2^2} + \ln\left(\frac{\sigma_2 \pi_1}{\sigma_1 \pi_2}\right).
\end{gathered}
\end{equation}

This quadratic equation may have two real roots. Among these, we select the solution that lies within the interval $(\min(\mu_1, \mu_2), \max(\mu_1, \mu_2))$ as the threshold $\tau$:
\[
\tau \in (\min(\mu_1, \mu_2), \max(\mu_1, \mu_2)).
\]
In rare cases where the two components have significant overlap or the EM fitting is degenerate, this solver may fail to yield a real root within the interval. This fallback ensures the method remains robust even when the ideal solution is not identifiable. In such cases, we define the threshold heuristically as the midpoint between the two means:
\begin{equation}
\tau = \frac{\mu_1 + \mu_2}{2}.
\end{equation}
Once the threshold $\tau$ is determined, binary labels are assigned to each spot based on whether the predicted score exceeds the threshold:
\begin{equation}
\hat{c}_i =
\begin{cases}
1, & \text{if } \hat{y}_i \ge \tau \quad \text{(cancer)} \\
0, & \text{otherwise} \quad \text{(healthy)}
\end{cases}.
\end{equation}

The proposed GMM-based thresholding is fully data-driven and adaptive to the distribution of each sample’s scores, providing a robust and interpretable approach, especially beneficial in weakly-supervised settings and datasets with heterogeneous annotation quality.

\section{Datasets}
\subsection{Dataset Descriptions}
As shown in Table~\ref{data}, we conducted extensive benchmark experiments on datasets from two types of cancer to demonstrate the accuracy and generalizability of SpaCRD:\par
\noindent \textbf{Human Breast Cancer}: The human breast cancer (HBC) datasets used in this study consist of eight HBC datasets generated from the older ST platform (denoted as STHBC\_A to STHBC\_H) \cite{her2st}, one HBC dataset generated from the 10X Visium platform (denoted as ViHBC) \cite{vihbc}, one human invasive ductal carcinoma (IDC) dataset generated from the 10X Visium platform (denoted as IDC) \cite{bayes}, and one HBC dataset generated from the Xenium platform (denoted as XeHBC) \cite{xenium}. The eight STHBC\_[A-H] datasets were derived from different oncology patients. For cross-samples CTR detection, we performed leave-one-out cross-validation on the eight STHBC\_[A-H] datasets. For cross-platforms\&batches CTR detection, the model was trained on the STHBC datasets and tested on the ViHBC, XeHBC, and IDC datasets.\par
\noindent \textbf{Human Colorectal Cancer}: To further demonstrate the generalizability of SpaCRD in CTR detection, we additionally conducted experiments on twelve colorectal cancer datasets generated from the 10X Visium platform (denoted as CRC\_[A1-E2, G1, G2]) \cite{crc}. The twelve CRC datasets represent samples from multiple tumor patients, where pairs like A1 and A2 correspond to one patient, B1 and B2 to another, and so on. For cross-sample CTR detection, we performed a patient-wise leave-one-out cross-validation on the twelve CRC datasets.\par
\noindent \textbf{Histology Image}: Each ST dataset is accompanied by a corresponding histology image, spatial coordinates of all spots relative to the image, the spot radius, and the pixel resolution of the histology image.\par
\subsection{Dataset Preprocessing}
All datasets were preprocessed through a standardized pipeline. Spots located outside tissue regions were removed based on the ``in\_tissue'' flag provided in the spatial transcriptomics metadata. Gene expression matrices were normalized to a total count of 10{,}000 per spot and log-transformed using the SCANPY package. Finally, the top 3{,}000 highly variable genes (HVGs) were selected using the Seurat v3 method implemented in SCANPY.\par
Since the XeHBC dataset is at single-cell resolution and thus differs from the spot-based ST data distribution, we performed an aggregation preprocessing step to make it comparable. Specifically, we aggregated cells based on the Visium spot size, diameter, and center-to-center distance by grouping all cells covered within each spot area:
\begin{equation}
    s_{i}=\sum_{k\in \Gamma _{i} }c_{k},
\end{equation}
where $s_{i}$ represents the gene expression of spot $S_{i}$, and $\Gamma_{i}$ represents the set of all cells covered by spot $S_{i}$. This process generated data comparable to the 10X Visium spatial resolution.

\begin{table*}[!t]
\centering
\renewcommand{\arraystretch}{1.5} 
\resizebox{1\textwidth}{!}{
\begin{tabular}{|l|l|l|l|c|}
\hline
Datasets & Tissue & Spots & Cancer Ratio & Platform \\ \hline
STHBC\_A & Breast cancer (Invasive cancer \& Cancer in situ) & 346 & 89.9\% & \multirow{8}{*}{ST} \\ \cline{1-4}
STHBC\_B & Breast cancer (Invasive cancer) & 295 & 22.0\% & \\ \cline{1-4}
STHBC\_C & Breast cancer (Invasive cancer) & 176 & 73.9\% & \\ \cline{1-4}
STHBC\_D & Breast cancer (Invasive cancer) & 306 & 48.1\% & \\ \cline{1-4}
STHBC\_E & Breast cancer (Invasive cancer) & 587 & 57.9\% & \\ \cline{1-4}
STHBC\_F & Breast cancer (Invasive cancer) & 691 & 85.7\% & \\ \cline{1-4}
STHBC\_G & Breast cancer (Invasive cancer \& Cancer in situ) & 467 & 34.3\% & \\ \cline{1-4}
STHBC\_H & Breast cancer (Invasive cancer \& Cancer in situ) & 613 & 33.1\% & \\ \hline
XeHBC    & Breast cancer (Ductal carcinoma in situ \& invasive carcinoma) & 4,050 & 34.9\% & Xenium \\ \hline
ViHBC    & Breast cancer (Lobular carcinoma in situ \& Invasive Carcinoma) & 3,798 & 61.9\% & \multirow{14}{*}{Visium} \\ \cline{1-4}
IDC      & Breast cancer (invasive carcinoma \& carcinoma in situ) & 4,727 & 59.9\% & \\ \cline{1-4}
CRC\_A1  & Colorectal cancer & 2,203 & 27.0\% & \\ \cline{1-4}
CRC\_A2  & Colorectal cancer & 2,385 & 31.2\% & \\ \cline{1-4}
CRC\_B1  & Colorectal cancer & 2,317 & 67.0\% & \\ \cline{1-4}
CRC\_B2  & Colorectal cancer & 1,803 & 70.9\% & \\ \cline{1-4}
CRC\_C1  & Colorectal cancer & 328 & 25.6\% & \\ \cline{1-4}
CRC\_C2  & Colorectal cancer & 1,048 & 42.3\% & \\ \cline{1-4}
CRC\_D1  & Colorectal cancer & 691 & 86.1\% & \\ \cline{1-4}
CRC\_D2  & Colorectal cancer & 1,219 & 91.2\% & \\ \cline{1-4}
CRC\_E1  & Colorectal cancer & 1,192 & 36.9\% & \\ \cline{1-4}
CRC\_E2  & Colorectal cancer & 387 & 54.0\% & \\ \cline{1-4}
CRC\_G1  & Colorectal cancer & 2,128 & 15.9\% & \\ \cline{1-4}
CRC\_G2  & Colorectal cancer & 1,691 & 20.5\% & \\ \hline
\end{tabular}}
\caption{Summary of all experimental datasets used in this study.}
\label{data}
\end{table*}

\section{Implementation Details}
\begin{figure*}[!h]
\centering
\includegraphics[width=1\textwidth]{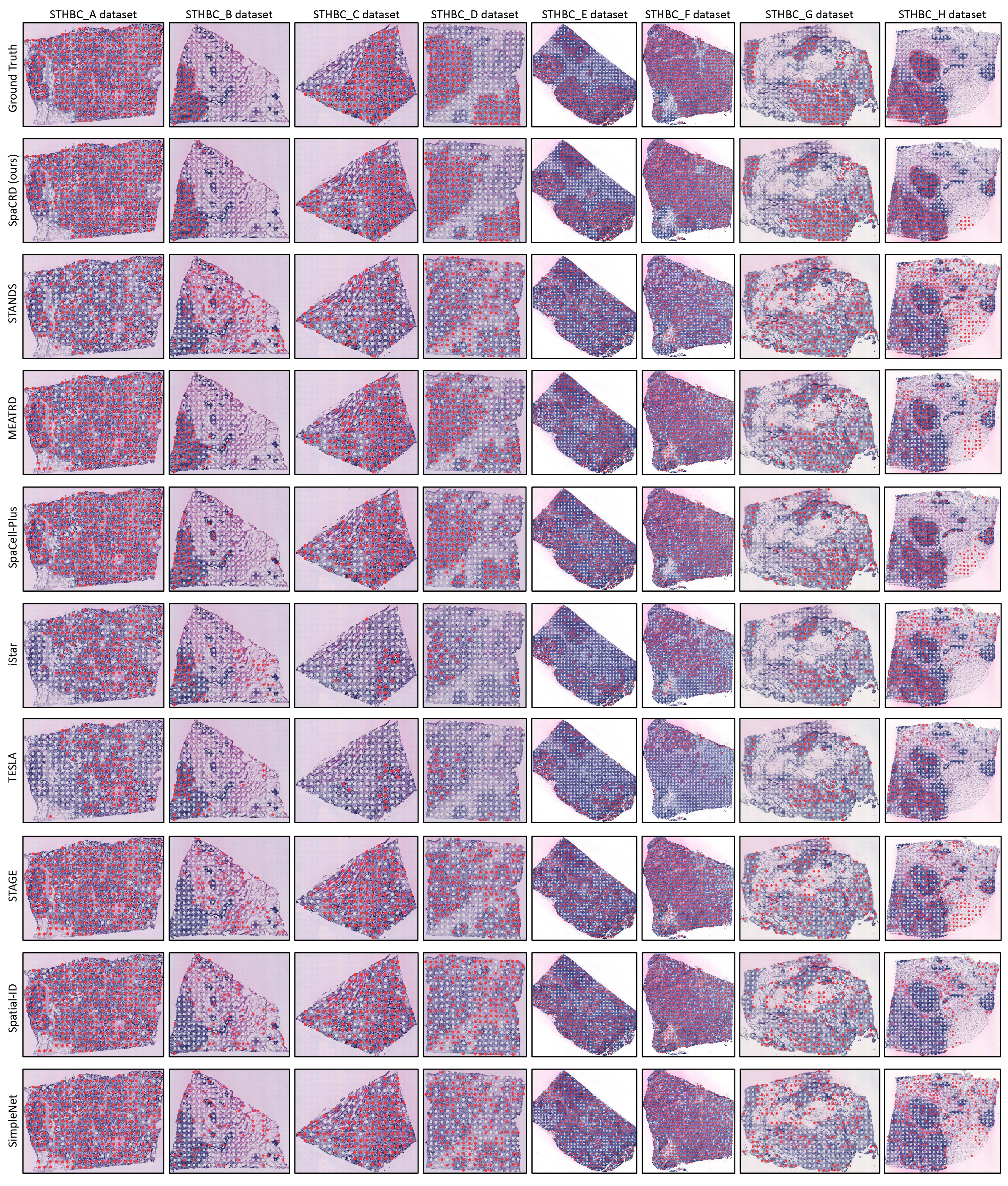}
\caption{ATR detection results of SpaCRD and other baselines on the eight STHBC datasets. Gray and red dots represent normal and cancerous spots, respectively.}
\label{figs1}
\end{figure*}

\begin{figure*}[t]
\centering
\includegraphics[width=1\textwidth]{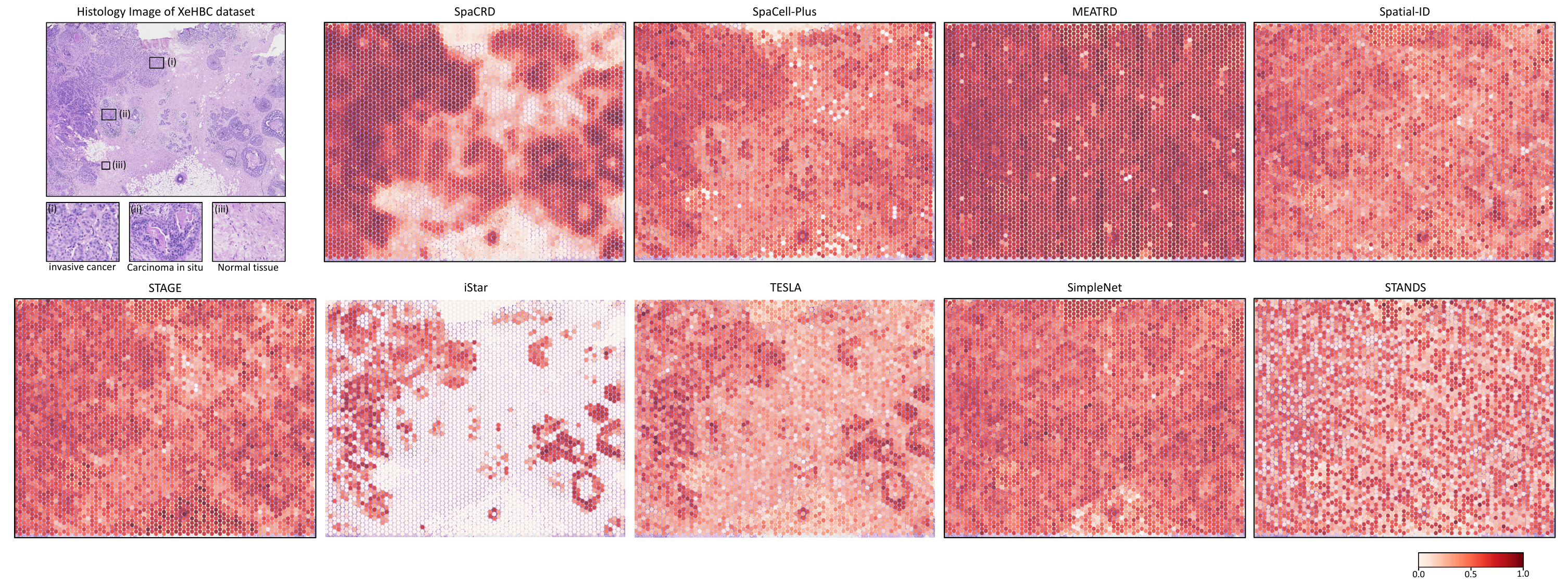}
\caption{Visualization of cancer likelihood scores predicted by SpaCRD and other baselines on the XeHBC dataset.}
\label{figs2}
\end{figure*}

\begin{figure*}[!h]
\centering
\includegraphics[width=1\textwidth]{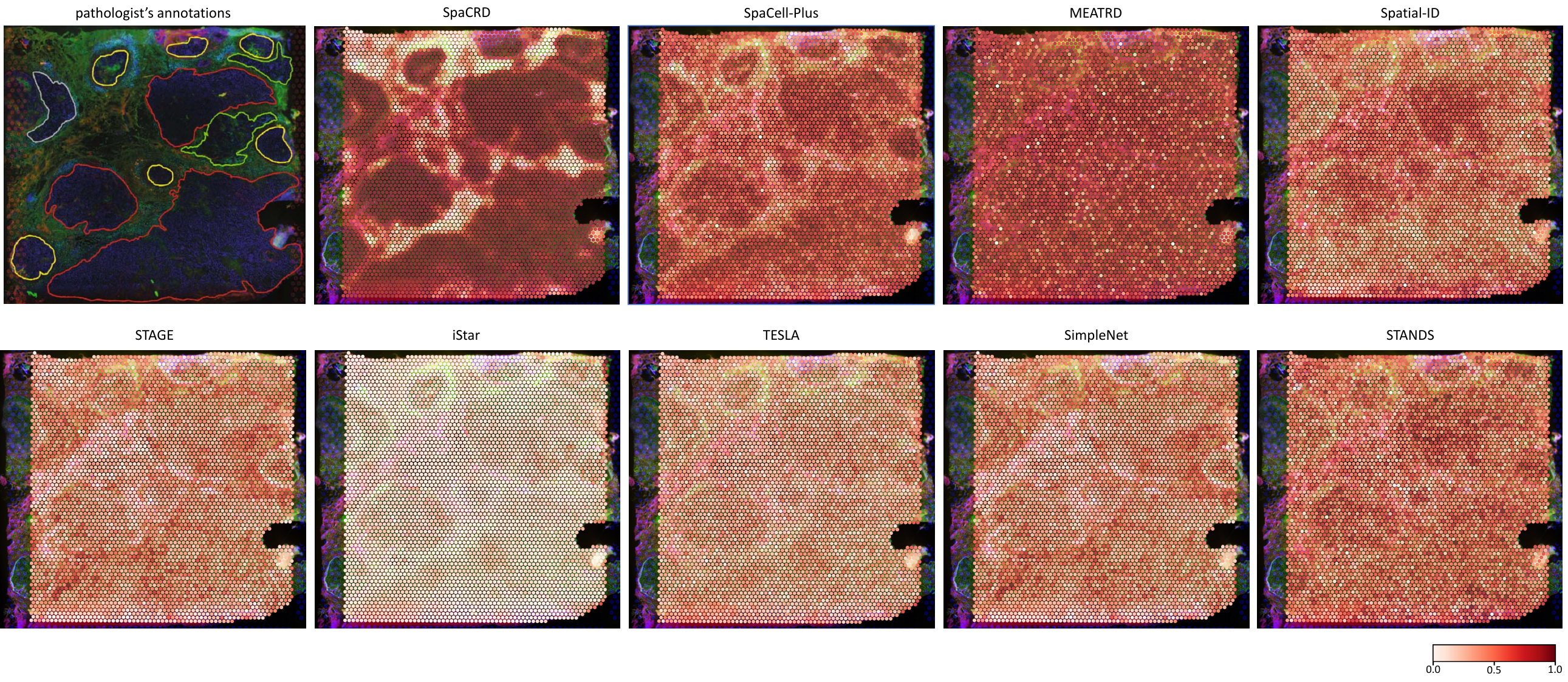}
\caption{Visualization of cancer likelihood scores predicted by SpaCRD and other baselines on the IDC dataset.}
\label{figs3}
\end{figure*}

We conducted all experiments using a single NVIDIA RTX 3090 GPU (24GB), with the development environment based on PyTorch 2.1.1 and Python 3.11.5. The three training stages were run for 100, 50, and 50 epochs, respectively, using the Adam optimizer with a uniform learning rate of 0.00001 across all stages. For contrastive learning, both the image and gene encoder comprise two hidden layers with 1024 and 512 units, followed by a projection layer that maps the features into a shared 512-dimensional latent space. Batch normalization, ReLU activation, and dropout with a rate of 0.2 are applied to each hidden layer. The BCA module consists of two asymmetric cross-attention layers followed by a linear fusion layer. Each cross-attention layer employs 8 attention heads with an embedding dimension of 512. The multimodal features from both directions are concatenated and projected through a linear fusion layer into a unified 512-dimensional representation, followed by LayerNorm. For RVAE module, the encoder consists of two fully connected layers with 256 and 128 units, each followed by a ReLU activation, and then two parallel linear layers that output the latent mean $\boldsymbol{\mu}$ and log-variance $\log \boldsymbol{\sigma}^2$, both of dimension 64. The decoder is symmetric to the encoder, comprising fully connected layers that reverse this transformation to reconstruct the input. Two learnable cluster centers are initialized from Gaussian distributions with means $-0.5$ and $+0.5$, respectively. Finally, the latent mean $\boldsymbol{\mu}$ and log-variance $\log \boldsymbol{\sigma}^2$ are concatenated and fed into a two-layer MLP-based classifier with a hidden size of 64 to predict cancer likelihood scores. For the bidirectional cross-attention mechanism, we select the top 6 nearest neighboring spots for each spot based on Euclidean distance to enable deep contextual information fusion.

\begin{figure*}[!t]
\centering
\includegraphics[width=1\textwidth]{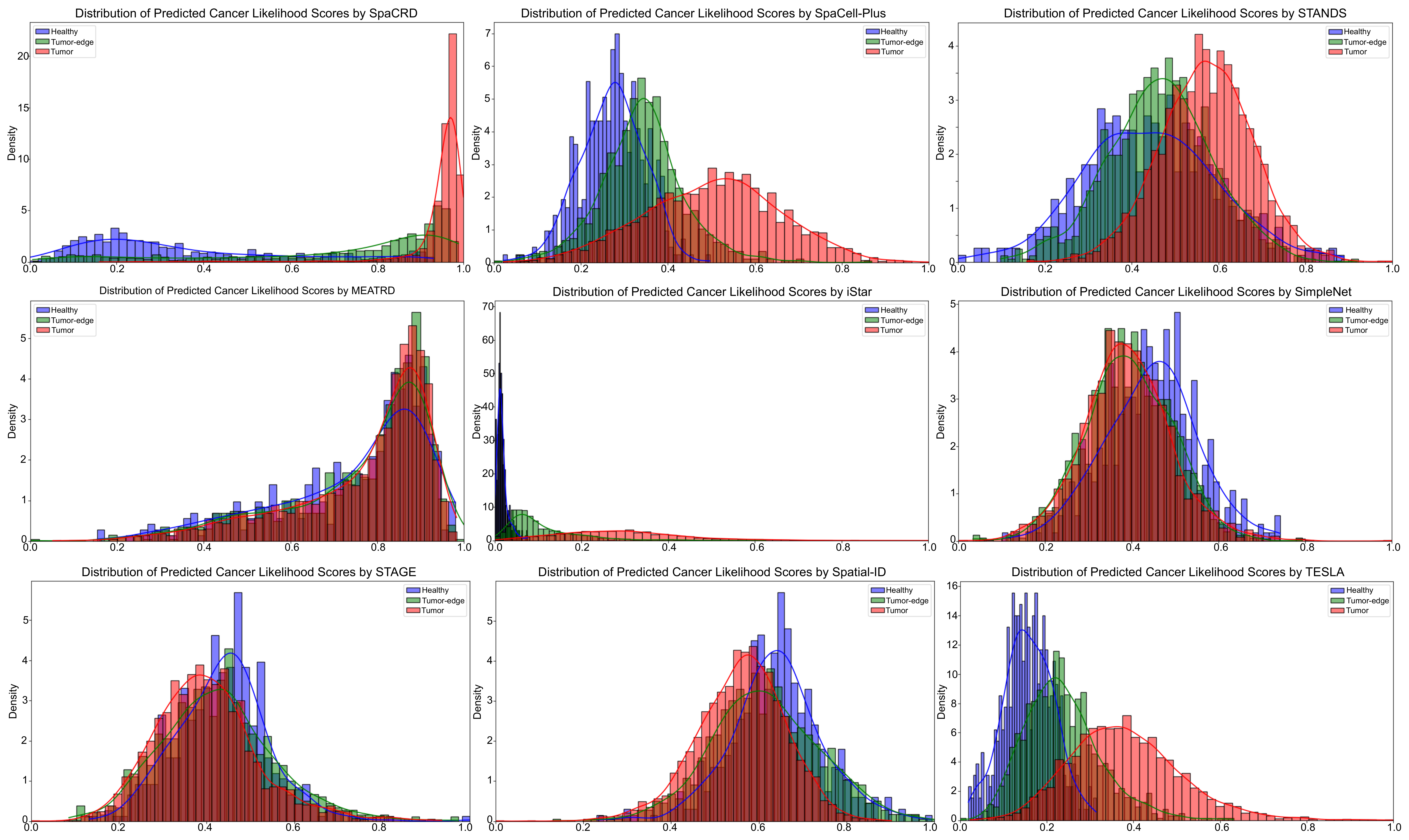}
\caption{Histograms of predicted cancer likelihood scores by SpaCRD and other baselines on the ViHBC dataset, grouped by three region types: healthy, tumor, and tumor boundary, based on ground truth.}
\label{figs5}
\end{figure*}

\section{Baseline methods}  
\textbf{MEATRD} identifies disease regions by learning to reconstruct multimodal features from normal tissues and detecting regions with high reconstruction errors. It first extracts visual features from $32\times 32$  histology patches using Mobile-UNet and encodes the corresponding gene expression profiles with a two-layer MLP. These dual-modal representations are then fused within 3-hop neighborhood subgraphs using a Masked Graph Dual-Attention Transformer. The resulting 128-dimensional feature is compressed into a 64-dimensional bottleneck, from which the histology image and gene expression are independently reconstructed using a ResNet and a GNN, respectively. MEATRD is trained on normal tissue samples, and disease regions are identified during inference based on reconstruction errors. In our implementation, to evaluate MEATRD on the STHBC\_[A-H], ViHBC, XeHBC, and IDC datasets for CTR detection, the model was trained on eight human healthy breast datasets, denoted as HHB\_[v03-v10]. For colorectal cancer datasets (CRC\_[A1–E2, G1, G2]), MEATRD was trained on two normal human colon (CRC\_F1 and CRC\_F2) datasets, and inference was performed separately on each CRC\_[A1–E2, G1, G2] dataset. All other settings and hyperparameters followed the original MEATRD implementation as described in the original paper.\par
\textbf{STANDS} formulates disease region detection as an anomaly detection task based on generative adversarial reconstruction. A GAN generator is trained on a normal reference dataset to reconstruct both gene expression and histology features. For each spot, a multimodal representation is constructed by fusing a gene embedding (obtained via GAT) and a histology embedding (extracted using ResNet-GAT) through a Transformer module, resulting in a 512-dimensional latent feature. The decoder then reconstructs both the original gene expression profile and the corresponding $112 \times 112$ histology patch. After training, the reconstruction error—measured as the cosine distance between the original and reconstructed features—is computed on the target dataset and used as the disease likelihood score. To evaluate STANDS in our experiments, the model was trained on the HHB\_v05, HHB\_v06, and HHB\_v07 datasets and tested on the STHBC\_[A–H], ViHBC, XeHBC, and IDC datasets following the original paper’s inference procedure. We followed the training data selection strategy as described in the original study, which uses these specific reference datasets, rather than the full range HHB\_[v03–v10]. For the CRC datasets, the training and inference datasets were the same as those used for MEATRD. All other settings and hyperparameters followed the original STANDS implementation as described in the original paper.\par
\textbf{SpaCell-Plus} is an extended version of SpaCell, designed to support cross-platform CTR detection by incorporating transfer learning. Specifically, it extracts features from $299 \times 299$ histology image patches using ResNet50 and processes them in parallel with the gene count matrix of each ST spot through two completely independent autoencoders. The resulting latent vectors of equal dimension are concatenated and fed into the same cancer likelihood discriminator as used in our method for prediction. While retaining the original architecture of SpaCell, SpaCell-Plus replaces the unsupervised training paradigm with a transfer learning strategy. To ensure a fair comparison, SpaCell-Plus was trained and tested on the same datasets as SpaCRD. \par
\textbf{iStar} primarily relies on expert-defined prior knowledge to perform CTR detection. It identifies cancer-epithelial cells by comparing predefined marker gene lists with the gene expression profiles in the ST data. The model architecture of iStar primarily consists of a pre-trained HIPT module for extracting hierarchical histology features, followed by a feedforward neural network for predicting super-resolution gene expression. To enable benchmark-compatible CTR detection, we aggregate the cancer-epithelial likelihood scores of all superpixels covered by each Visium-scale spot, based on the spot size and spatial distribution pattern defined by the Visium platform. Since iStar is entirely based on prior knowledge, it does not require training data and can be directly applied to the test datasets for inference.\par
\textbf{TESLA} relies primarily on expert-defined prior knowledge for tumor region detection. It infers cancer regions by comparing predefined tumor marker gene lists with a super-resolution gene expression map imputed from low-resolution ST data and H\&E-stained histology images. The model architecture comprises two main components: (i) a superpixel-level gene expression imputation module that integrates spatial proximity and histological similarity, and (ii) a lightweight CNN that segments tumor regions based on a meta-gene image concatenated with the grayscale histology. To ensure compatibility with benchmark evaluations, cancer likelihood scores from all superpixels within each Visium-scale spot are aggregated according to the spot geometry and spatial layout defined by the Visium platform. As TESLA is fully prior-driven, it does not require any training data and can be directly applied to test datasets for inference. \par
\textbf{STAGE} was originally designed to impute gene expression at unmeasured locations in spatial transcriptomics by integrating spatial coordinates and gene expression through a spatially supervised autoencoder. Specifically, the model encodes gene expression profiles into latent representations, which are trained to reconstruct the original inputs under the supervision of spot coordinates. In our adaptation for CTR detection, we replace the spatial supervision with binary labels indicating cancerous or non-cancerous spots, thereby modeling the latent representation as a cancer/healthy embedding. We further adopt a transfer learning strategy, enabling the model to generalize across datasets. \par

\begin{figure*}[!t]
\centering
\includegraphics[width=1\textwidth]{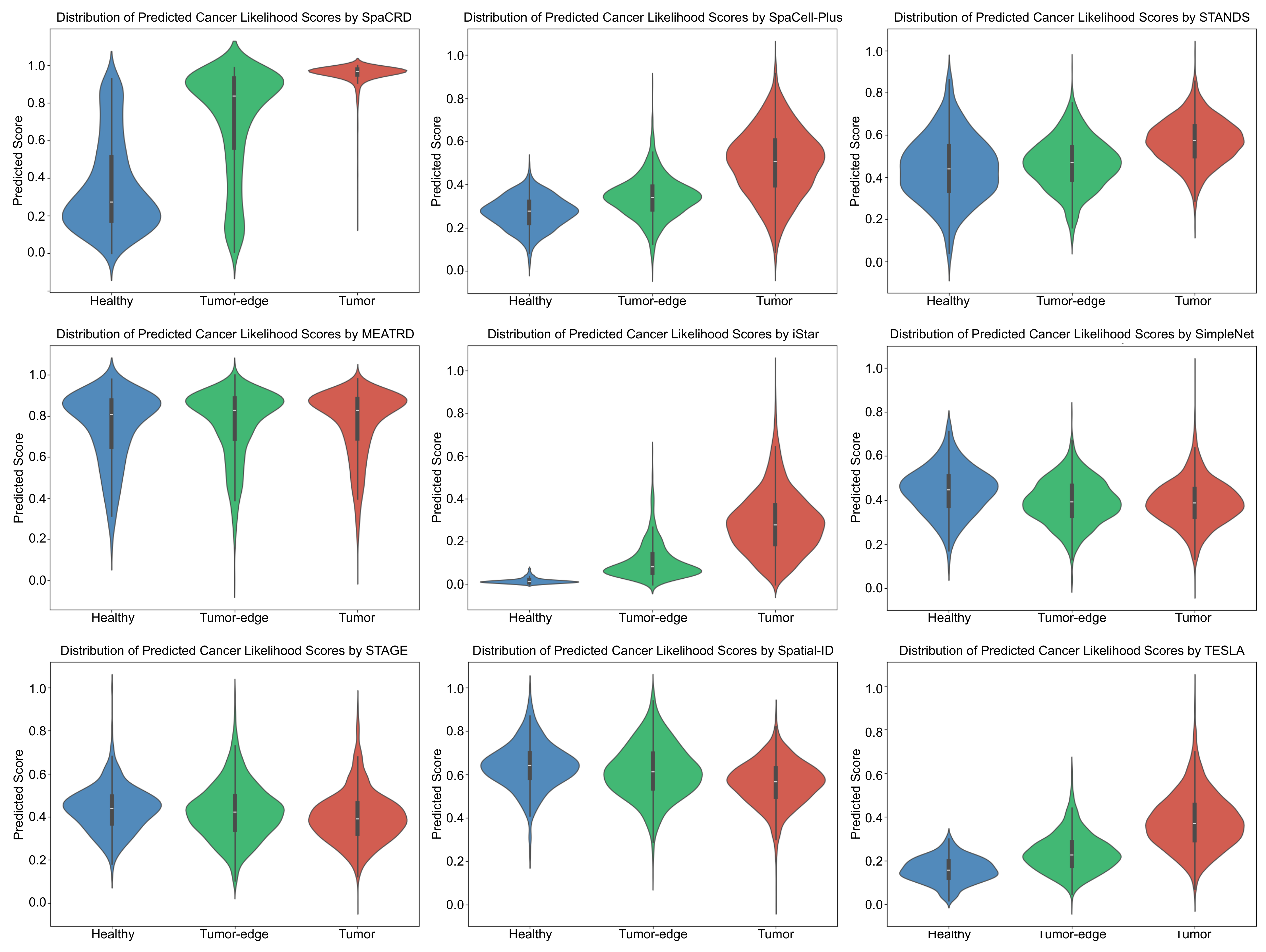}
\caption{Violin plots of predicted cancer likelihood scores by SpaCRD and other baselines on the ViHBC dataset, grouped by three region types: healthy, tumor, and tumor boundary, based on ground truth.}
\label{figs6}
\end{figure*}

\textbf{Spatial-ID} is a spatial clustering method designed to distinguish between cancerous and healthy spots. It first trains a deep neural network (DNN) using annotated scRNA-seq data to generate cell type probability distributions as pseudo-labels. Then, a spatial adjacency graph is constructed, and a variational graph autoencoder (VGAE) is employed to integrate gene expression with spatial coordinates. A self-supervised learning scheme is used to optimize a cell-type classifier, which ultimately identifies ambiguous spots with low maximum assignment probabilities. Since Spatial-ID relies on scRNA-seq data for pretraining, we adopt the publicly available pretrained model and skip the training-from-scratch stage in our implementation. \par
\textbf{SimpleNet} is a lightweight anomaly (i.e., cancer) detection framework consisting of a pretrained feature extractor, a single fully connected feature adapter, a Gaussian noise-based anomaly generator, and a two-layer MLP discriminator. It introduces random noise into embedded visual features and trains a hyperplane-based discriminative factor to separate pseudo-anomalies from normal samples. For fair comparison, SimpleNet is evaluated using the same dataset partitioning as MEATRD. \par

\section{Evaluation Metrics}
To comprehensively evaluate the performance of SpaCRD in CTR detection, we employ four standard classification metrics: Area Under the Receiver Operating Characteristic curve (AUC), Average Precision (AP), F1-score, and Kolmogorov-Smirnov (KS) distance.

\textbf{Area Under the ROC Curve (AUC)} measures the area under the receiver operating characteristic curve, which plots the true positive rate against the false positive rate across all classification thresholds. It is mathematically defined as:
\begin{equation}
\text{AUC} = \int_0^1 \text{TPR}(t) \, d\text{FPR}(t),
\end{equation}
where $\text{TPR}(t) = \frac{\text{TP}(t)}{\text{TP}(t) + \text{FN}(t)}$ and $\text{FPR}(t) = \frac{\text{FP}(t)}{\text{FP}(t) + \text{TN}(t)}$ represent the true positive rate and false positive rate at threshold $t$, respectively. AUC ranges from 0 to 1, with higher values indicating superior discriminative capability across all operating points.

\textbf{Average Precision (AP)} summarizes the precision-recall curve by computing the area under it, providing a single scalar that captures performance across all recall levels:
\begin{equation}
\text{AP} = \int_0^1 P(r) \, dr = \sum_{k=1}^{n} P_k \Delta R_k,
\end{equation}
where $P_k$ and $\Delta R_k = R_k - R_{k-1}$ denote the precision and recall increment at the $k$-th threshold, respectively. AP is particularly valuable for imbalanced datasets where the minority class (positive samples) is of primary interest.

\textbf{F1-score} represents the harmonic mean of precision and recall, providing a balanced measure that accounts for both false positives and false negatives:
\begin{equation}
\text{F1} = \frac{2 \cdot \text{Precision} \cdot \text{Recall}}{\text{Precision} + \text{Recall}} = \frac{2\text{TP}}{2\text{TP} + \text{FP} + \text{FN}},
\end{equation}
where $\text{Precision} = \frac{\text{TP}}{\text{TP} + \text{FP}}$ and $\text{Recall} = \frac{\text{TP}}{\text{TP} + \text{FN}}$. To ensure a fair comparison, the F1-score was computed using a threshold that matches the actual proportion of true positives in the ground truth.

\textbf{Kolmogorov-Smirnov (KS) distance} quantifies the difference between two distributions by finding the largest gap between their cumulative distribution functions. A KS distance near 1 signifies strong separation between the distributions. It is mathematically defined as:
\begin{equation}
D_{\text{KS}} = \sup_x |F(x) - G(x)|
\end{equation}
where \(\sup_x\) denotes the supremum over all values of \(x\).

\section{Further Results}  

\begin{table}[ht]
\centering
\renewcommand{\arraystretch}{1.1} 
\resizebox{1\columnwidth}{!}{
\begin{tabular}{lcccc}
\toprule
Method & Params & Training Time (s) & Inference Time (s) & Memory Usage (MB) \\
\midrule
SpaCRD (Ours) & 8.70M & 33.86 & 1.15 & 535.68 \\
MEATRD & 48.61M & 1378.70 & 13.35 & 8401.39 \\
STANDS & 68.19M & 986.77 & 12.89 & 18426.91 \\
SpaCell-Plus & 1.72M & 44.25 & 0.06 & 94.14 \\
iStar & 2.41M & 222.03 & 11.98 & 325.15 \\
TESLA & 5.68M & 5790.68 & 18.71 & 7619.64 \\
STAGE & 5.21M & 770.06 & 0.12 & 60.36 \\
SimpleNet & 75.63M & 1.66 & 10.27 & 1.28 \\
Spatial-ID & 4.12M & 967.18 & 1.16 & 0.37 \\
\bottomrule
\end{tabular}}
\caption{Comparison of computational cost on the ViHBC dataset among all methods.}
\label{runtime}
\end{table}
\subsection{Visualization of CTR Detection Results}
We further demonstrate the clear superiority of SpaCRD over all baseline methods by visualizing the CTR detection results on the eight STHBC\_[A–H] datasets. As shown in Figure~\ref{figs1}, SpaCRD achieves the highest consistency with ground-truth CTR annotations in most datasets. Its superiority is especially pronounced in challenging cases involving multiple spatially disconnected CTR, such as STHBC\_D, STHBC\_E, and STHBC\_G.\par
Subsequently, we visualized the cancer likelihood scores predicted by SpaCRD and all baseline methods on the XeHBC and IDC datasets to further assess their performance on cross-platform\&batch CTR detection. As shown in Figure~\ref{figs2} and \ref{figs3}, SpaCRD not only effectively separates normal and cancerous tissue regions, but also captures differences in cancer severity based on the magnitude of the predicted likelihood scores (e.g., distinguishing between invasive carcinoma and carcinoma in situ). In contrast, although other methods can to some extent distinguish cancerous regions from normal tissue, their predicted boundaries are often overly blurred, and the likelihood scores show insufficient separation among invasive carcinoma, carcinoma in situ, and normal regions. Such limitations are inadequate for CTR detection tasks that require high accuracy.\par

\begin{table*}[ht]
\centering
\renewcommand{\arraystretch}{1.3} 
\resizebox{1\textwidth}{!}{
\begin{tabular}{l|ccccccccc}
\toprule
\multicolumn{1}{l|}{\multirow{2}{*}{Training data}} & \multicolumn{3}{c}{ViHBC}& \multicolumn{3}{c}{XeHBC} & \multicolumn{3}{c}{IDC}\\ \cline{2-10} 
\multicolumn{1}{c|}{}                         & \multicolumn{1}{c}{AUC} & \multicolumn{1}{c}{AP} & \multicolumn{1}{c}{F1} & \multicolumn{1}{c}{AUC} & \multicolumn{1}{c}{AP} & \multicolumn{1}{c}{F1}& \multicolumn{1}{c}{AUC} & \multicolumn{1}{c}{AP} & \multicolumn{1}{c}{F1} \\ \hline
C (176 spots)   & 0.861\scriptsize{$\pm$0.029}& 0.882\scriptsize{$\pm$0.035}& 0.817\scriptsize{$\pm$0.018}& 0.875\scriptsize{$\pm$0.022}& 0.749\scriptsize{$\pm$0.047}& 0.731\scriptsize{$\pm$0.038}& 0.779\scriptsize{$\pm$0.034}& 0.783\scriptsize{$\pm$0.36}& 0.757\scriptsize{$\pm$0.41}\\
D (306 spots)   & 0.888\scriptsize{$\pm$0.016}& 0.913\scriptsize{$\pm$0.016}& 0.863\scriptsize{$\pm$0.010}& 0.911\scriptsize{$\pm$0.016}& 0.820\scriptsize{$\pm$0.037}& 0.759\scriptsize{$\pm$0.025}& 0.867\scriptsize{$\pm$0.021}& 0.886\scriptsize{$\pm$0.26}& 0.833\scriptsize{$\pm$0.36}\\ \hline
Full (3,481 spots) & 0.900\scriptsize{$\pm$0.029}& 0.930\scriptsize{$\pm$0.037}& 0.867\scriptsize{$\pm$0.014}& 0.931\scriptsize{$\pm$0.008}& 0.859\scriptsize{$\pm$0.021}& 0.795\scriptsize{$\pm$0.013}& 0.891\scriptsize{$\pm$0.006}& 0.914\scriptsize{$\pm$0.006}& 0.854\scriptsize{$\pm$0.008}\\
\bottomrule
\end{tabular}}
\caption{Small-sample training analysis results on the ViHBC, XeHBC, and IDC datasets.}
\label{small_study}
\end{table*}

\subsection{Distribution Analysis of Cancer Likelihood Scores Across Tissue Regions}
To further evaluate the performance of SpaCRD, we performed a distribution analysis of cancer likelihood scores on the ViHBC dataset, which contains manual annotations dividing the tissue into three distinct regions: healthy, tumor, and tumor-edge. We analyzed the cancer likelihood scores of spots within each annotated region to assess the ability of different methods to distinguish these biologically meaningful areas. As shown in Figure~\ref{figs5}, SpaCRD effectively distinguishes between healthy and tumor regions, with the predicted cancer likelihood scores exhibiting well-separated distributions (Mean: healthy=0.35, tumor-edge=0.70, tumor=0.94; Mode: healthy=0.19, tumor-edge=0.91, tumor=0.97). SpaCRD predicts tumor-edge regions with likelihood scores closer to tumor regions, reflecting that these edge regions are transitional zones that may warrant special attention due to potential pathological changes. In contrast, other methods such as SpaCell-Plus (Mean: healthy=0.27, tumor-edge=0.34, tumor=0.50; Mode: healthy=0.28, tumor-edge=0.34, tumor=0.53) and iStar (Mean: healthy=0.02, tumor-edge=0.11, tumor=0.28; Mode: healthy=0.01, tumor-edge=0.06, tumor=0.28) show smaller separations in predicted scores across the three regions, suggesting weaker contrast between healthy, tumor-edge, and tumor spots. Furthermore, as shown in Figure~\ref{figs5}, the predicted scores for spots across the three regions are highly overlapping in other baseline methods, making it difficult to clearly distinguish between healthy, tumor-edge, and tumor regions. To further highlight the advantages of SpaCRD, we visualized the distributions of the three regions using violin plots. As shown in Figure~\ref{figs6}, SpaCRD (top-left) demonstrates a clear separation among the predicted cancer likelihood scores for healthy, tumor-edge, and tumor regions. The healthy spots mostly have low scores, tumor-edge spots show intermediate scores closer to tumor regions, and tumor spots have high scores approaching 1.0. This distribution aligns well with the biological progression from healthy to tumor states, especially highlighting the transitional nature of tumor-edge regions.

Finally, we evaluated the distributional divergence between predicted cancer likelihood scores in healthy and tumor regions across the eight STHBC datasets using the KS distance. As shown in Figure 4 of the main text, SpaCRD exhibited the strongest separation between healthy and tumor regions across most STHBC datasets (Median: SpaCRD=0.754, SpaCell-Plus=0.494, MEATRD=0.348, TESLA=0.366, STAGE=0.273, SimpleNet=0.264, iStar=0.354, Spatial-ID=0.186, STANDS=0.127). Overall, these results demonstrate the superior capability of SpaCRD to accurately reflect the underlying biological differences between healthy and tumor spots, which is critical for precise CTR detection.

\subsection{Sensitivity Analysis}
We conducted a sensitivity analysis on three key components: the directional weighting coefficient $\alpha$ for contrastive loss in modality-alignment representation learning, the category-regularized KL divergence loss weight $\beta$, and the multimodal fused loss weight $\gamma$ in the cancer likelihood discriminator.
Table 5 of the main text shows the average model performance over five independent runs across eleven breast cancer datasets.
We observed that setting $\alpha = 0.5$ yields the best performance. This result highlights the benefit of jointly modeling bidirectional modality alignment. However, the model remains relatively robust when using $\alpha = 0$ or $\alpha = 1$, with only minor performance differences.
For the category-regularized KL divergence loss, we observed that setting $\beta = 0.5$ achieves strong performance. Increasing $\beta$ to 1.0 only slightly decreases the performance, whereas reducing it to 0.1 leads to a notable drop. This indicates that incorporating a moderate level of class-aware regularization is essential for downstream cancer region identification.
Setting $\gamma$ to 0 results in a substantial performance degradation, as omitting the fused loss enables the cancer likelihood discriminator to adversely affect the latent space, causing overfitting.
Finally, we found that selecting 6 neighboring spots in the BCA module achieves optimal performance, consistent with the hexagonal arrangement of ST data. Although performance slightly decreases when selecting 4 or 8 neighbors, it remains stable, demonstrating the robustness of our model to the choice of neighborhood size.

\subsection{Empirical Efficiency Analysis}
We compared the empirical efficiency analysis of SpaCRD with all baseline methods. As shown in Table~\ref{runtime}, all models were trained on eight STHBC datasets and tested on ViHBC dataset. SpaCRD uses a moderate number of parameters (8.70M) and achieves much faster training (33.86 s) and inference times (1.15 s) than larger models like MEATRD and STANDS, while consuming significantly less memory (535.68 MB). Compared to lightweight models such as SpaCell-Plus and STAGE, SpaCRD maintains a good balance between runtime efficiency and resource usage. These results indicate that SpaCRD maintains an acceptable computational cost.

\subsection{Small-Sample Training Analysis}
We conducted a small-sample training analysis on the eleven breast cancer datasets. In our main experiments, the model was trained on all eight STHBC datasets (3,481 spots) and evaluated on three  test datasets (ViHBC, XeHBC, and IDC), each containing between 3,798 and 4,727 spots. To further investigate the effect of training data scale, we conducted additional experiments with significantly reduced training samples. Specifically, we trained our model on much smaller subsets—176 spots from the STHBC\_C dataset and 306 spots from the STHBC\_D dataset. As shown in Table~\ref{small_study}, although performance declined slightly under these conditions, the drop was not substantial, demonstrating that our model is robust and does not heavily rely on large-scale training data.

\end{document}